\documentclass[a4paper,fleqn,review]{cas-dc}

\usepackage[authoryear]{natbib}

\usepackage{amsmath,amssymb,amsthm}
\usepackage{epsfig}
\usepackage{pifont}
\newcommand{\cmark}{\ding{51}}
\newcommand{\xmark}{\ding{55}}
\usepackage{subfig}
\usepackage[ruled,vlined,algosection,linesnumbered]{algorithm2e}
\usepackage{lineno}
\usepackage{algorithmic}
\usepackage{xcolor}

\def\tsc#1{\csdef{#1}{\textsc{\lowercase{#1}}\xspace}}
\tsc{WGM}
\tsc{QE}

\begin{document}
\let\WriteBookmarks\relax
\def\floatpagepagefraction{1}
\def\textpagefraction{.001}

\shorttitle{PolyGNN}    

\shortauthors{Z. Chen~et al.}  

\title[mode = title]{PolyGNN: Polyhedron-based Graph Neural Network for 3D Building Reconstruction from Point Clouds}

\let\printorcid\relax 

\author[1]{Zhaiyu Chen}
\ead{zhaiyu.chen@tum.de}

\author[2]{Yilei Shi}
\ead{yilei.shi@tum.de}

\author[3]{Liangliang Nan}
\ead{liangliang.nan@tudelft.nl}

\author[1]{Zhitong Xiong}
\ead{zhitong.xiong@tum.de}

\author[1,4]{Xiao Xiang Zhu}
\ead{xiaoxiang.zhu@tum.de}

\cormark[1]

\affiliation[1]{organization={Chair of Data Science in Earth Observation, Technical University of Munich},
            postcode={80333},
            city={Munich},
            country={Germany}}
            
\affiliation[2]{organization={School of Engineering and Design, Technical University of Munich},
            postcode={80333},
            city={Munich},
            country={Germany}}

\affiliation[3]{organization={Urban Data Science, Delft University of Technology},
            postcode={2628 BL},
            city={Delft},
            country={The Netherlands}}
            
\affiliation[4]{organization={Munich Center for Machine Learning},
            postcode={80333},
            city={Munich},
            country={Germany}}

\cortext[1]{Corresponding author}


\begin{abstract}
We present PolyGNN, a polyhedron-based graph neural network for 3D building reconstruction from point clouds. PolyGNN learns to assemble primitives obtained by polyhedral decomposition via graph node classification, achieving a watertight and compact reconstruction. To effectively represent arbitrary-shaped polyhedra in the neural network, we propose a skeleton-based sampling strategy to generate polyhedron-wise queries. These queries are then incorporated with inter-polyhedron adjacency to enhance the classification. PolyGNN is end-to-end optimizable and is designed to accommodate variable-size input points, polyhedra, and queries with an index-driven batching technique. To address the abstraction gap between existing city-building models and the underlying instances, and provide a fair evaluation of the proposed method, we develop our method on a large-scale synthetic dataset with well-defined ground truths of polyhedral labels. We further conduct a transferability analysis across cities and on real-world point clouds. Both qualitative and quantitative results demonstrate the effectiveness of our method, particularly its efficiency for large-scale reconstructions. The source code and data are available at \url{https://github.com/chenzhaiyu/polygnn}.
\end{abstract}




\maketitle

\section{Introduction}

Three-dimensional (3D) building models constitute an important infrastructure in shaping digital twin cities, facilitating a broad range of applications including urban planning, energy demand estimation, and environmental analysis~\citep{biljecki2015applications}. Therefore, efficient reconstruction of high-quality 3D building models is crucial for understanding an urban environment and has been a long-standing challenge.

Most reconstruction methods are dedicated to detailed surfaces represented by dense triangles~\citep{kazhdan2013screened,erler2020points2surf}, irrespective of the ubiquitous piecewise planarity in the built environment. Alternatively, a compact polygonal representation with sparse parameters can adequately capture the geometry of urban buildings. 
To reconstruct compact polygonal building models, three categories of methods are commonly employed in practice. Constrained reconstruction methods~\citep{zhou20102,li2016manhattan} represent buildings with pre-defined templates or specific topologies. However, the limited variety of available templates or topologies constrains the expressiveness of these methods. Geometric simplification methods~\citep{bouzas2020structure,li2021feature} aim to obtain compact surfaces by simplifying dense triangle ones. These techniques, however, necessitate an input model that is precise in both its geometry and topology to ensure a faithful approximation. Primitive assembly methods~\citep{nan2017polyfit,huang2022city3d} produce polygonal surface models by pursuing an optimal assembly of a collection of geometric primitives. However, these methods often entail the use of handcrafted features and thus possess limited representational capacity.

Despite successes in various other applications, learning-based solutions for compact building modeling remain largely unexplored. Notably, Points2Poly~\citep{chen2022points2poly} stands out as a pioneering effort with the primitive assembly strategy. This method utilizes an implicit representation to learn building occupancy, followed by a Markov random field (MRF) to enhance compactness. However, it learns occupancy independently of the primitive-induced hypothesis, leading to inefficiencies in large-scale applications.

In this paper, we present PolyGNN, a polyhedron-based graph neural network for reconstructing building models from point clouds. PolyGNN leverages the decomposition of a building's ambient space into a set of polyhedra as strong priors. It learns to assemble the polyhedra to achieve a watertight and compact reconstruction, framed as end-to-end graph node classification. The neural network can be efficiently optimized, enabling reconstruction at scale.

\begin{figure}[t]
  \centering
  \centerline{\epsfig{figure=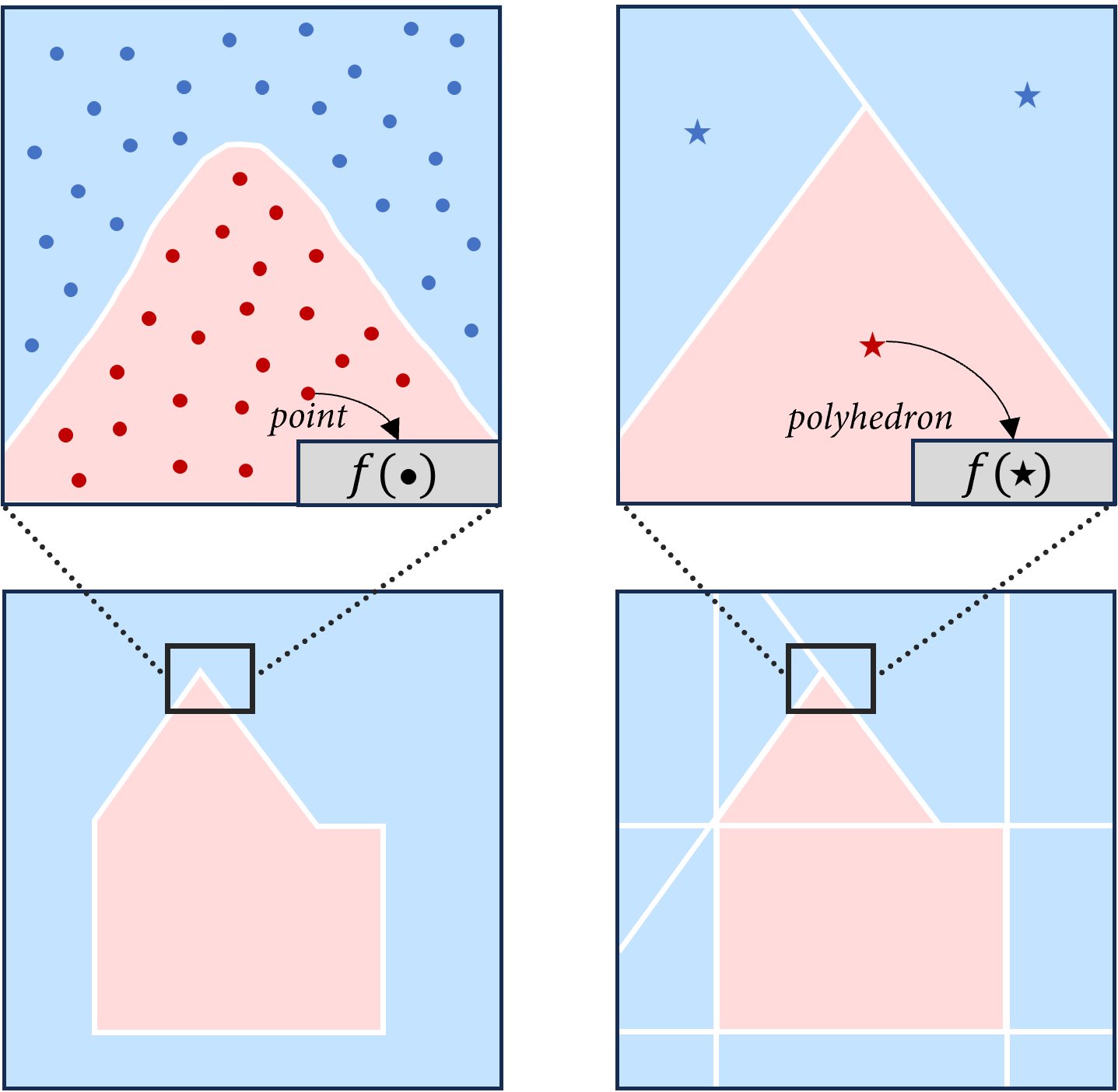,width=0.9\linewidth}}
\caption{Instead of learning a continuous function $f\left( \bullet \right)$ underpinned by exhaustive queries with traditional deep implicit fields, PolyGNN learns a piecewise planar occupancy function $f\left( \bigstar \right)$ supported by polyhedral decomposition.}
\label{fig:difference}
\end{figure}

Our key innovation entails integrating occupancy estimation with the polyhedral decomposition through primitive assembly. As illustrated in \autoref{fig:difference}, instead of learning a continuous function with traditional deep implicit fields, we opt for learning a piecewise planar occupancy function from the decomposition. There, one challenge lies in consistently representing the heterogeneous geometry of arbitrary-shaped polyhedra. To address this, we propose an efficient skeleton-based strategy to sample a set of representative points within the polyhedron as queries. These queries, conditioned on the latent building shape code, collectively characterize the building occupancy. Moreover, PolyGNN is designed to accommodate variable-size input points, polyhedra, and queries using an index-driven batching technique.

We observe that existing 3D city models offer abstract representations of real-world buildings, often lacking geometric details. Thus, employing existing mesh models as ground truths is inherently inadequate due to systematic errors. To facilitate supervised learning for PolyGNN, we resort to creating a large-scale synthetic dataset comprised of simulated airborne LiDAR point clouds and corresponding building models. This dataset enables reliable one-to-one mapping between the two sources, effectively addressing the potential abstraction gap. Subsequently, we assess the transferability of our method across cities and on real-world data.

The main contributions of this paper are summarized as follows:
\begin{itemize}
    \item We present PolyGNN, a polyhedron-based graph neural network for reconstructing compact polygonal building models from point clouds.
    \item We introduce a skeleton-based sampling strategy that efficiently represents arbitrary-shaped polyhedra within the neural network.
    \item We demonstrate the transferability of PolyGNN on cross-city synthetic point clouds and a real-world airborne LiDAR point cloud.
\end{itemize}

\section{Related work}

In this section, we discuss three categories of methods used for polygonal building model reconstruction: constrained reconstruction, geometric simplification, and primitive assembly. Subsequently, we introduce a line of relevant works in neural implicit representation, from which we draw inspiration for our work.

\subsection{Constrained reconstruction}

Constrained reconstruction methods reduce the complexity of reconstruction by incorporating constraints into the solution space, employing pre-defined templates or specific disk topologies.

With a library of roof model templates, \citet{henn2013model} applied estimation methods to derive best-fitting models from sparse LiDAR point clouds. The widely adopted Manhattan-world assumption restricts the orientation of building surfaces in three orthogonal directions, representing buildings with axis-aligned polycubes. \citet{vanegas2012automatic} provided a solution to reconstruct Manhattan-world building masses from LiDAR scans. \citet{li2016fitting} and \citet{li2016manhattan} further extended the solution with integer programming and an MRF, respectively. \citet{suveg2004reconstruction} proposed to integrate aerial image analysis with GIS footprints, and formulated reconstruction as a multilevel hypothesis generation and verification scheme.

Another common constraint is restricting output surfaces to specific disk topologies. Satellite data offers prospects for global building models which often lack the necessary level of detail (LoD), and thus are primarily limited to 3D models at LoD1~\citep{zhu2022,sun2022large}. The 2.5D view-dependent representation can generate building roofs with vertical walls connecting them from LiDAR measurements~\citep{zhou20102}. \citet{peters2022automated} employed footprint partitioning and extrusion for an automated building reconstruction in both LoD1 and LoD2. \citet{huang2022city3d} proposed an LoD2 building reconstruction approach with integer programming. Furthermore, \citet{xiong2014graph} and \citet{xiong2015flexible} exploited roof topology graphs for reconstructing LoD2 buildings from predefined primitives. \citet{chen2017topologically} proposed a three-stage method for LoD2 modeling that consists of primitive clustering, boundary representation, and geometric modeling. \citet{kelly2017bigsur} proposed a data fusion technique for structured urban reconstruction from coarse meshes, street-view imagery, and GIS footprints.

The constrained reconstruction approaches simplify the reconstruction and are thus efficient to implement. However, they only apply to specific domains as the limited variety of templates or topologies restricts the expressiveness of these methods. Our reconstruction method, instead, does not rely on templates or specific topologies, thus remaining generic.

\subsection{Geometric simplification}

Another category of approaches generates compact 3D models through the simplification of existing dense triangle meshes typically obtained from photogrammetric surface reconstruction methods or learning-based alternatives. We refer to \citet{berger2017survey} for the former, and \citet{sulzer2023survey} for the latter.

\citet{garland1997surface} introduced a mesh simplification method based on iterative vertex contraction using quadratic error metrics. \citet{salinas2015structure} further incorporated planar proxies detected from a pre-processing analysis to preserve piecewise planar structures. The variational shape approximation technique~\citep{cohen2004variational} optimizes a set of geometric proxies to construct an approximating polygonal mesh.

Several methods specifically focus on the simplification of urban scenes. \citet{verdie2015lod} proposed an approach that incorporates various LoD configurations through classification, abstraction, and reconstruction. \citet{bouzas2020structure} incorporated a structure graph to encode planar primitives and formulated geometric simplification as mesh polygonization. \citet{li2021feature} utilized planar regions to constrain edge collapse operations, achieving feature-preserving building mesh simplification. \citet{gao2022low} proposed a three-stage approach involving carving visual hulls to generate low-poly building meshes.

While geometric simplification methods hold promise for compact building modeling, they often require high-quality input meshes for faithful surface approximation. In practice, these methods may introduce uncertainty and additional burdens without guaranteeing accuracy. In contrast, our proposed method can directly generate a compact polygonal mesh without approximating an intermediate.

\subsection{Primitive assembly}

Primitive assembly methods produce compact polygonal surface models by optimizing the assembly of a set of geometric primitives.
In practice, planar primitives can be extracted with RANSAC~\citep{schnabel2007efficient}, region growing~\citep{rabbani2006segmentation}, or other optimization methods~\citep{yu2022finding,li2019supervised,li2023surface}. High-quality primitives are critical to primitive assembly methods.

Connectivity-based approaches~\citep{chen2008architectural,van2011shape,schindler2011classification} address the assembly by extracting proper geometric primitives from an adjacency graph built on planar shapes. While efficient in analyzing the graph, these methods are sensitive to the quality of the graph. Linkage errors contaminating the connectivity can compromise the reconstruction. A hybrid strategy proposed by \cite{lafarge2013surface} and \citet{holzmann2018semantically} represented high-confidence areas by polygons and more complex regions by dense triangles. \citet{arikan2013snap} presented an interactive optimization-based snapping solution, which requires labor-intensive human involvement in handling complex structures.

Slicing-based approaches are more robust to imperfect data with the hypothesis-and-selection strategy. With the primitives, these approaches~\citep{chauve2010robust,mura2016piecewise,nan2017polyfit,bauchet2020kinetic} partition the 3D space into polyhedral cells by extending the primitives to supporting planes, transforming the reconstruction into a labeling problem where the polyhedral cells are labeled as either inside or outside the shape or equivalently by labeling other primitives. \citet{li2021relation} extended PolyFit~\citep{nan2017polyfit} to leverage the inter-relation of the primitives for procedural modeling. \citet{huang2022city3d} further extended PolyFit by introducing a new energy term to encourage roof preferences and two additional hard constraints to ensure correct topology and enhance detail recovery. \citet{fang2020connect} proposed a hybrid approach for reconstructing 3D objects by successively connecting and slicing planes identified from 3D data. Further, \citet{xie2021combined} proposed an approach combining rule-based and hypothesis-based strategies. 

Our method inherits primitive assembly while realizing the selection by a graph neural network with the polyhedra from convex decomposition. The selection is therefore completely data-driven and requires no handcrafted features.

\subsection{Implicit neural representation}

Recent advances in deep implicit fields have revealed their potential for 3D surface reconstruction~\citep{park2019deepsdf,peng2020convolutional,erler2020points2surf}, and also specifically for buildings~\citep{stucker2022implicity}. The crux of these methods is to learn a continuous function to map the input, such as a point cloud, to a scalar field. The surface can then be extracted using iso-surfacing techniques like Marching Cubes~\citep{lorensen1987marching}. To learn a more regularized field, \citet{rella2022neural} and \citet{yang2023neural} both proposed to learn the displacements from queries towards the surface and model shapes as vector fields. However, these methods still require iso-surfacing to extract the final surfaces. Although iso-surfacing is effective in extracting smooth surfaces, it struggles to preserve sharp features, and introduces discretization errors. Consequently, deep implicit fields alone are unsuitable for reconstructing compact polygonal models. Notably, while \citet{chen2020bsp} and \citet{deng2020cvxnet} both employed implicit fields for reconstructing convex shapes obtained via binary space partitioning, transferring them for compact building reconstruction from point clouds remains elusive.

Points2Poly~\citep{chen2022points2poly} is a pioneering learning-based effort for polygonal building reconstruction. The key enabler is a learned implicit function that indicates the occupancy of a building, followed by an MRF for a favorable geometric complexity. By its design, Points2Poly is composed of two separate parts, hence cannot be optimized end-to-end. The occupancy learning is agnostic of the hypothesis. This limits its exploitation of deep features, and in turn, limits its efficiency. The prohibitive complexity hinders its application at scale. In contrast, our method directly learns to classify the polyhedra with an end-to-end neural architecture, underpinning great efficiency.

\section{Methodology}

\subsection{Overview}

\begin{figure}[t]
    \centering
    \subfloat[Candidates]{\includegraphics[width=0.28\linewidth, keepaspectratio]{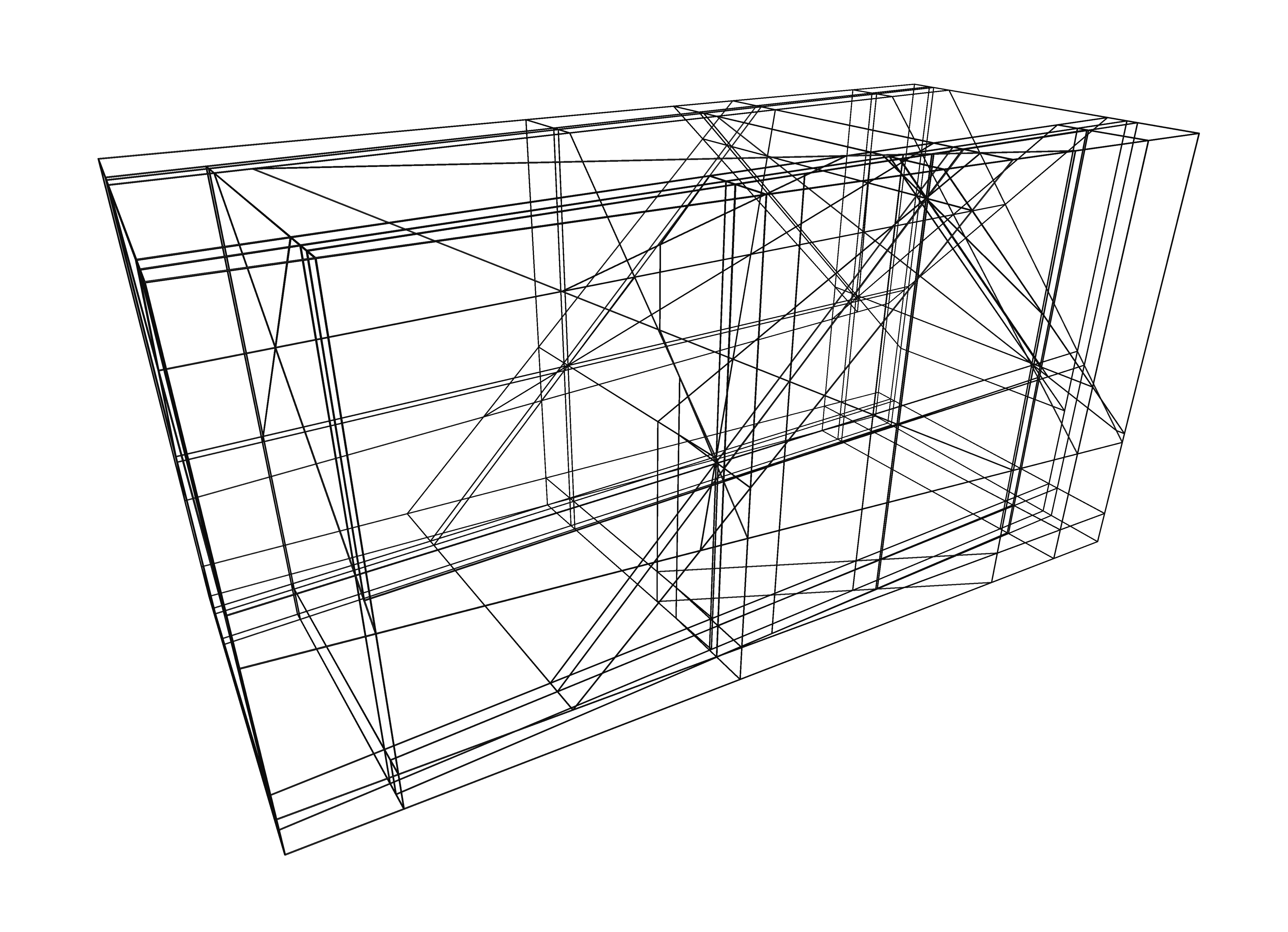}}
    \hspace{.5em}
    \subfloat[Classified]{\includegraphics[width=0.28\linewidth, keepaspectratio]{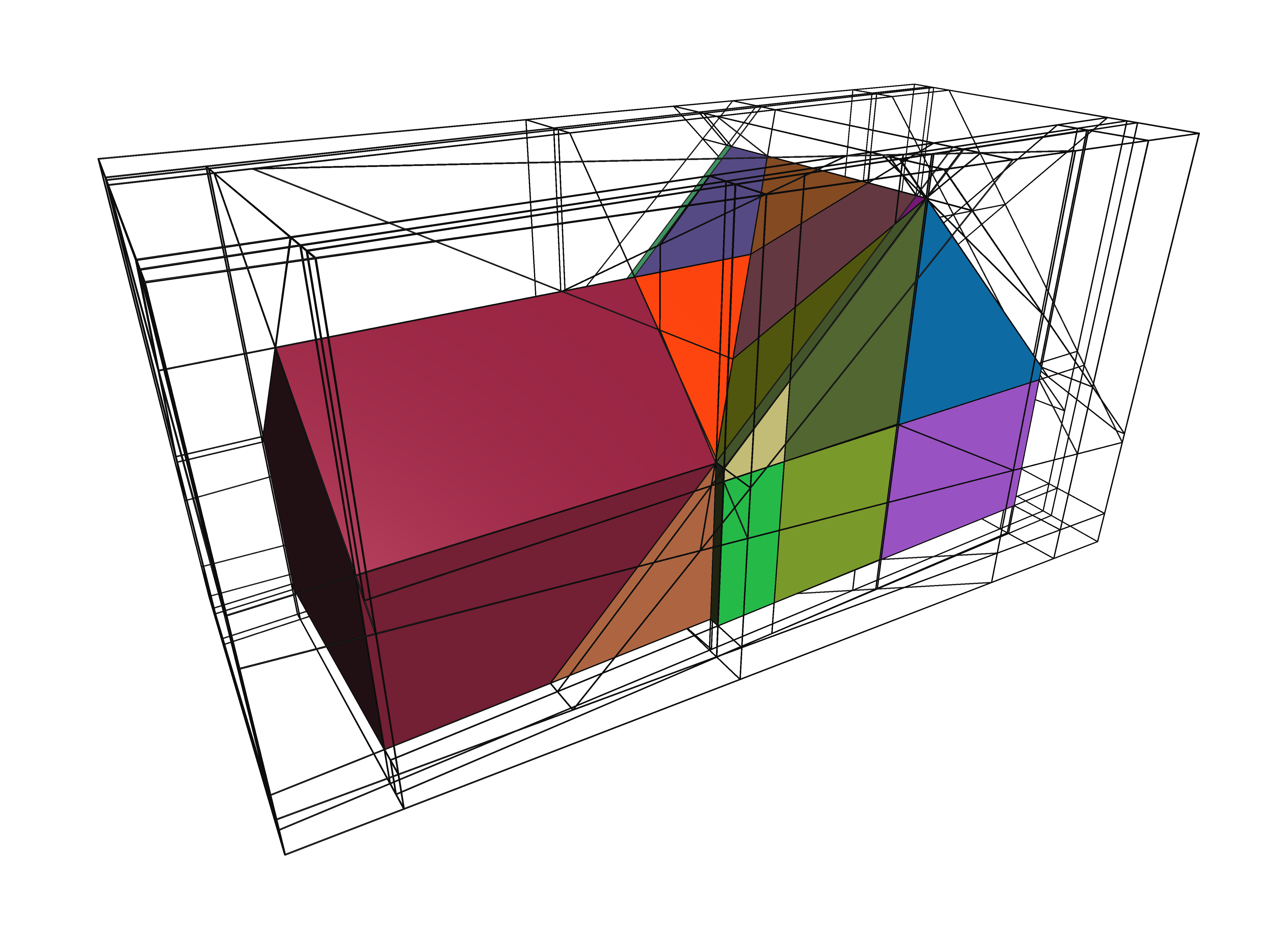}}
    \hspace{.5em}
    \subfloat[Surface]{\includegraphics[width=0.28\linewidth, keepaspectratio]{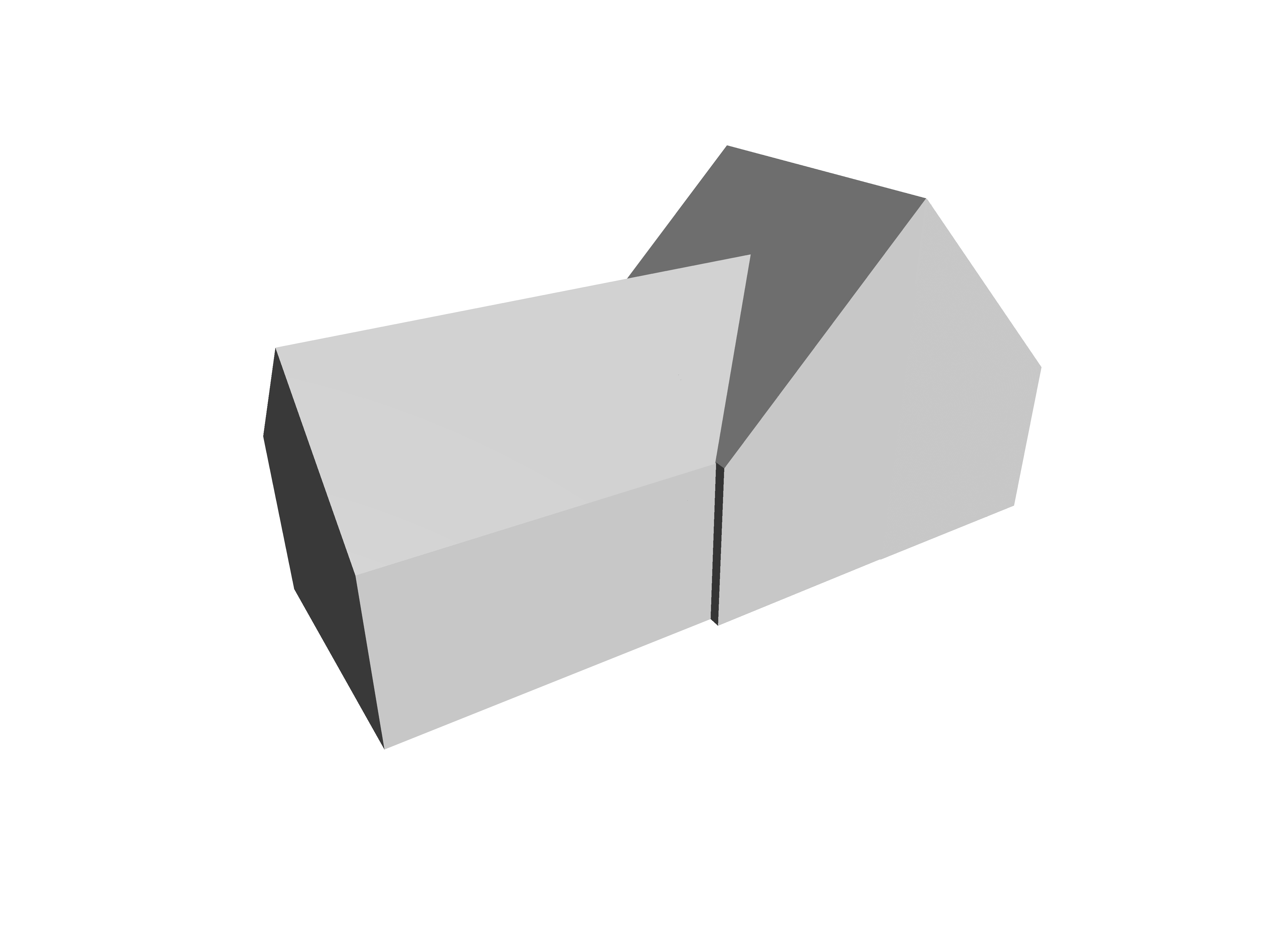}} \\
    \subfloat[Candidates]{\includegraphics[width=0.28\linewidth, keepaspectratio]{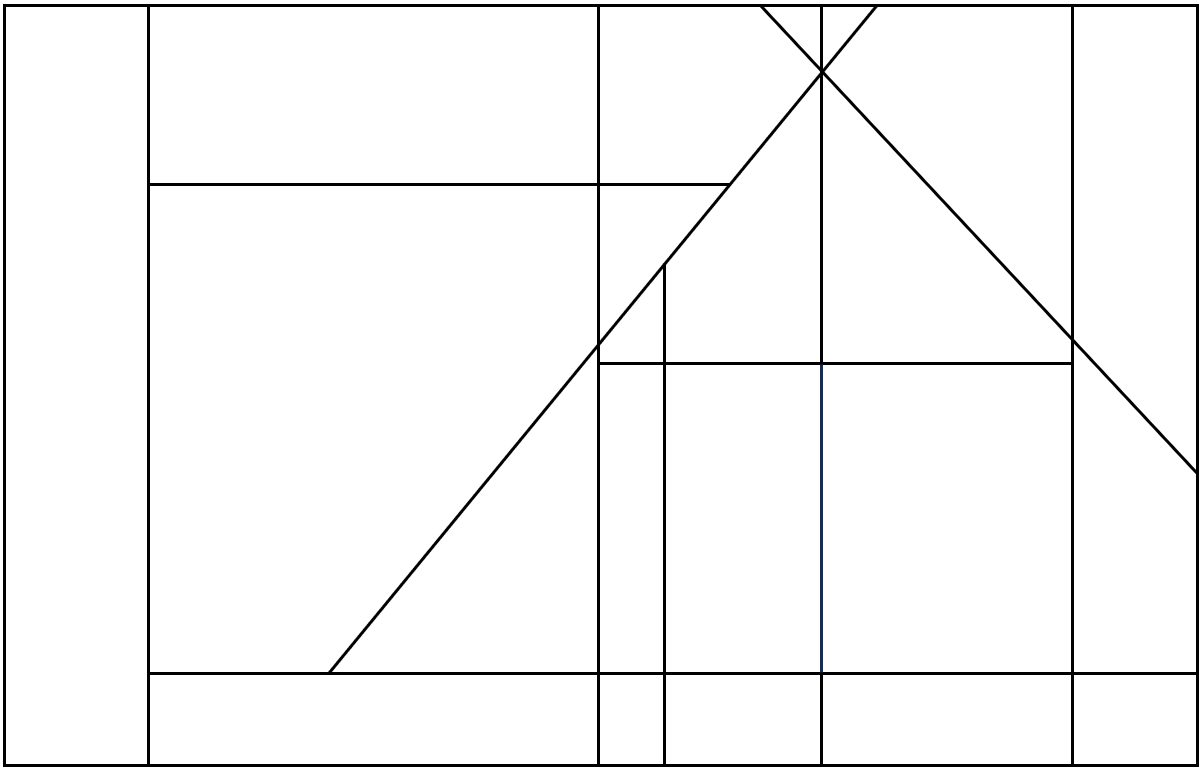}}
    \hspace{.5em}
    \subfloat[Classified]{\includegraphics[width=0.28\linewidth, keepaspectratio]{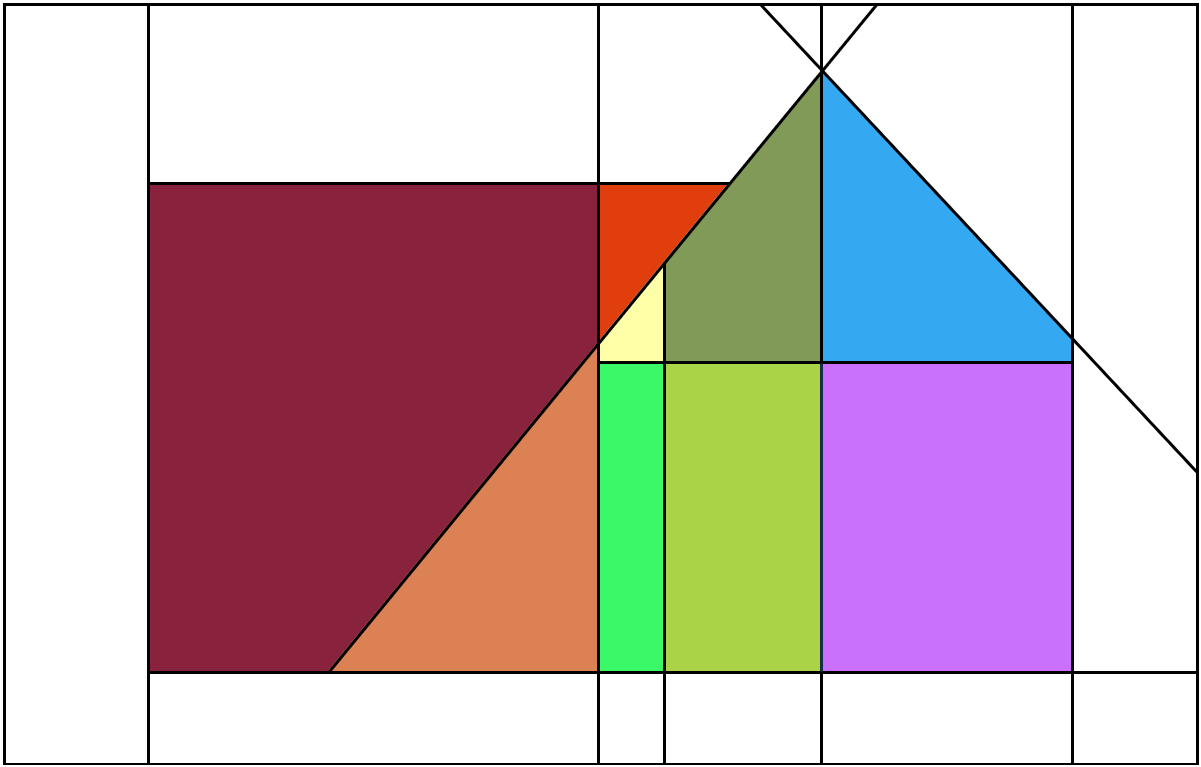}}
    \hspace{.5em}
    \subfloat[Surface]{\includegraphics[width=0.28\linewidth, keepaspectratio]{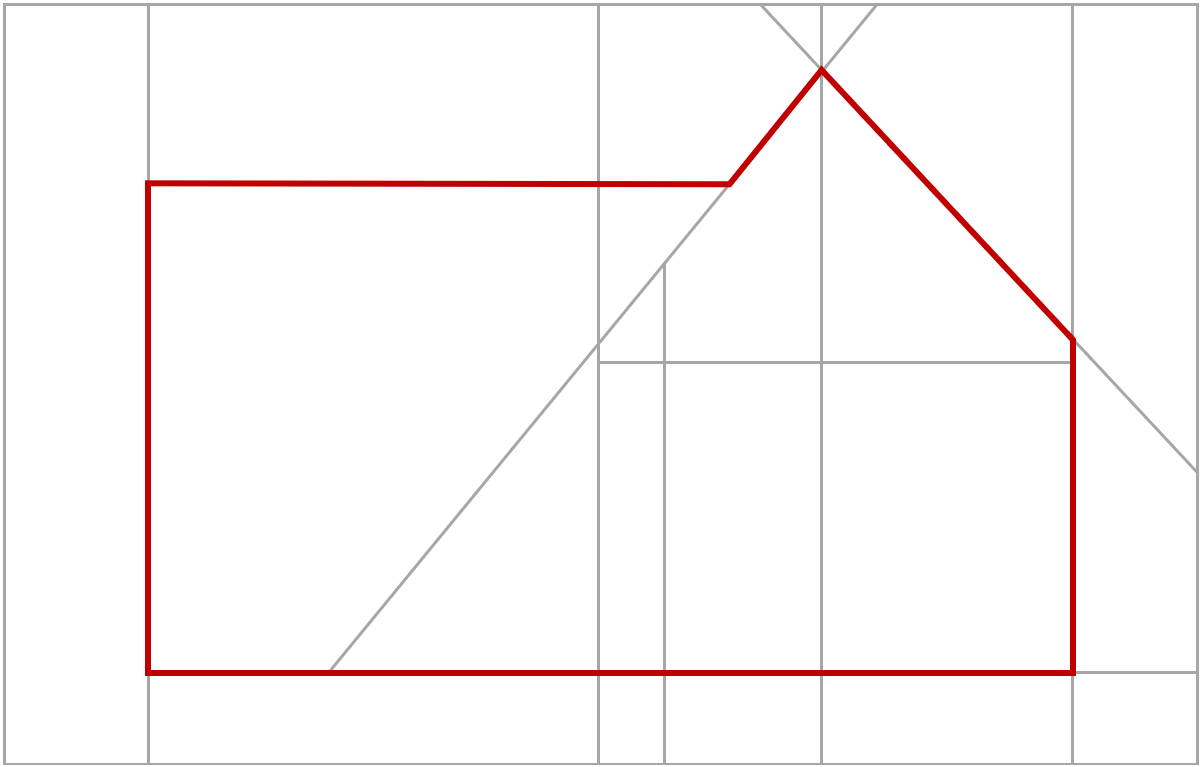}}
    \caption{Reconstruction by polyhedra classification. Candidate polyhedra (a) are generated by polyhedral decomposition and are classified by PolyGNN into \emph{interior} ones and \emph{exterior} ones (b). The surface (c) is extracted in between pairs of polyhedra of different classes. (d) (e) (f) are illustrations of 2D cross sections of (a) (b) (c), respectively.
    }
    \label{fig:partition}
\end{figure}

We formulate building reconstruction as a graph node classification problem. As illustrated in \autoref{fig:partition}, we first decompose the ambient space of a building into a cell complex of candidate polyhedra following binary space partitioning. We then represent the cell complex as a graph structure and classify the polyhedral nodes into two classes: \emph{interior} and \emph{exterior}. Finally, the building surface model can be extracted as the boundary between the two classes of polyhedra.

\begin{figure*}[htb]
  \centering
  \centerline{\epsfig{figure=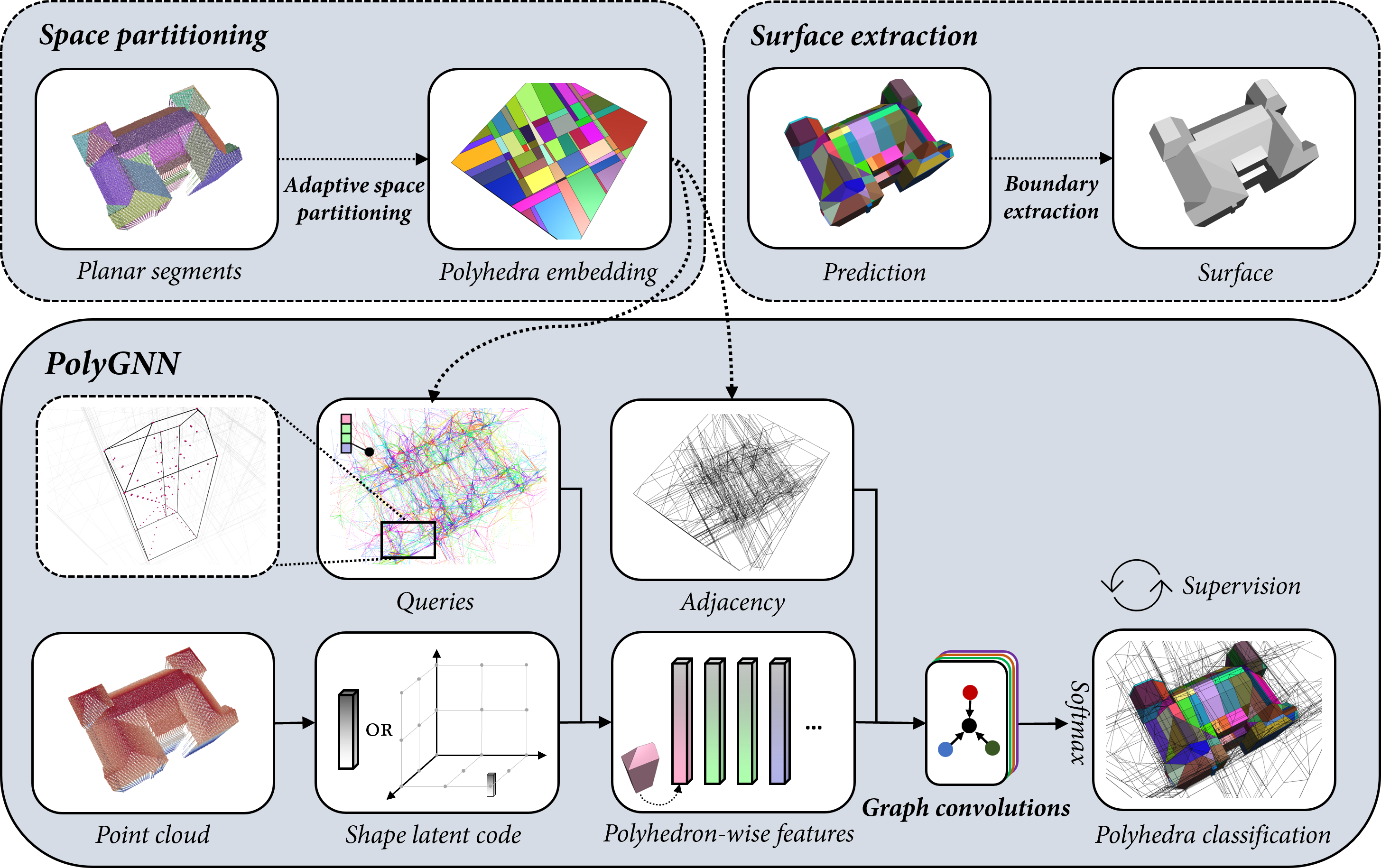,width=0.95\linewidth}}
\caption{Architecture of PolyGNN. Given an input point cloud, a graph topology is constructed from polyhedral decomposition, with polyhedra being graph nodes. Node features are formed by conditioning polyhedron-wise queries on a shape latent code. With the encoded node features and inter-polyhedron adjacency, graph nodes are classified for building occupancy estimation. Space partitioning and surface extraction are two external solvers.
}
\label{fig:architecture}
\end{figure*}

The above procedure can be formulated as follows. Given an unordered point set $\mathcal{X} = \{x_1, x_2, ..., x_n\}$ with $x_i \in \mathbb{R}^3$ as input, we first decompose the ambient space into an undirected graph embedding $\mathcal{G} = \left ( \mathcal{V}, \mathcal{E} \mid \mathcal{X} \right )$, where $\mathcal{V} = \{v_1, v_2, ..., v_m\}$ and $\mathcal{E} \subseteq \mathcal{V} \times \mathcal{V}$ represent non-overlapping convex polyhedra and their edges, respectively. $\mathcal{G}$ serves as a volumetric embedding, from which we seek an appropriate subset of $\mathcal{V}$ to align with the occupancy of the underlying building instance.
The surface reconstruction is therefore transformed into an assignment problem which we address with a graph neural network $\tilde{f}$:
\begin{equation}
    \tilde{f} \approx {f}(\mathcal{V} \mid \mathcal{X}, \mathcal{E}) = Y,
\label{eq:classification}
\end{equation}

where $Y = \{y_1, y_2, ..., y_m\} \subseteq \{0, 1\}$. \autoref{fig:architecture} illustrates the architecture of PolyGNN for solving the graph node classification problem, which consists of two stages:
\begin{itemize}
    \item \textbf{Polyhedral graph encoding}. A graph structure is constructed from polyhedral decomposition, with polyhedra being graph nodes. Node features are formed by conditioning polyhedron-wise queries on a shape latent code that encodes the occupancy of the building.
    \item \textbf{Polyhedral graph node classification}. With the encoded node features and inter-polyhedron adjacency, graph nodes (i.e., polyhedra) are classified for building occupancy estimation.
\end{itemize}
In the following, we elaborate on the two main components of the network.

\subsection{Polyhedral graph encoding}

\subsubsection{Polyhedral graph construction}
\label{subsec:graph}
\begin{figure}[htb]
  \centering \centerline{\epsfig{figure=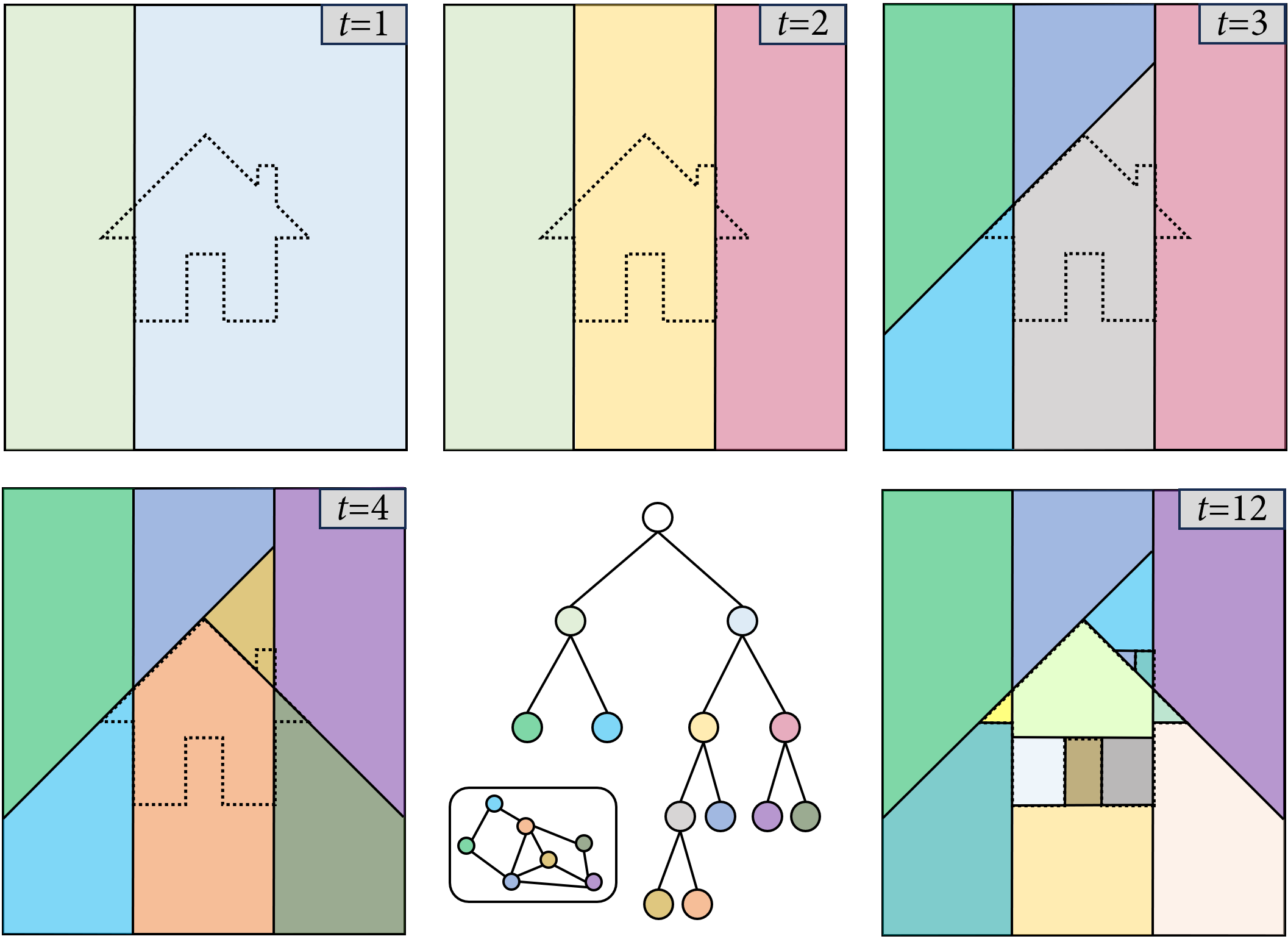,width=0.95\linewidth}}
\caption{Illustration of adaptive binary space partitioning. During partitioning, a binary tree is dynamically constructed to analyze inter-polyhedron adjacency. $t$ denotes iteration.}
\label{fig:adaptive}
\end{figure}

We adopt the adaptive binary space partitioning approach introduced by \citet{chen2022points2poly}. We first identify a set of planar primitives from the input point cloud that comprises the building, and subsequently partition the ambient 3D space to generate a linear cell complex of non-overlapping polyhedra that complies with the primitives. 
Given our focus on primitive assembly, we assume that high-quality planar primitives have been extracted.
As illustrated in \autoref{fig:adaptive}, vertical primitives and primitives with larger areas are given higher priority. The tessellation is spatially adaptive therefore being efficient and respective to the building’s geometry. The partitioning also involves the construction of a binary tree, which records the hierarchical information of polyhedra and their inter-polyhedron adjacency.

\subsubsection{Point cloud encoding}

By adaptive binary space partitioning, the polyhedral embedding $\mathcal{G}$ is produced in the Euclidean space, from which we seek an appropriate subset of $\mathcal{V}$ to align with the occupancy of the underlying building instance. In addition, we obtain the shape latent code $\mathbf{z}$ embedded in the feature space, via a graph neural network.

We transform $\mathcal{X}$ into a neural feature representation, a process which, in principle, can be achieved through any point cloud encoder. We choose a lightweight plain encoder and a more modern convolutional encoder, and demonstrate the performance of PolyGNN with the two encoders.

With the plain encoder, point features are encoded with layers of dynamic edge convolutions~\citep{wang2019dynamic}:
\begin{equation}
    g_{x_i}^{\left( 1 \right)} = \sum_{x_j\in\mathcal{N} \left ( x_i \right )} h_{\Theta} \left( \left[h_{x_i}, h_{x_j}, e_{x_i,x_j} \right] \right),
\end{equation}
where $g_{x_i}^{\left( 1 \right)}$ is the feature representation of point $x_i$. $\mathcal{N} \left ( x_i \right )$ denotes the set of neighboring points of $x_i$ in the K-nearest-neighbor graph constructed from the input point cloud. $e_{x_i,x_j}$ is the edge feature between $x_i$ and $x_j$. $h_{x_i}$ and $h_{x_j}$ are the features of points $x_i$ and $x_j$, respectively. $h_{\Theta}$ is a multi-layer perceptron (MLP) that maps the concatenated input to a new feature space. Features from $L$ layers are further concatenated and aggregated by a max pooling operator, followed by another MLP $\gamma_{\Theta}$ to form the global latent code denoted as $\mathbf{z}^{\left( 1 \right)}$:
\begin{equation}
\label{eq:latent_plain}
\mathbf{z}^{\left( 1 \right)} = \gamma_{\Theta} \max_{i=1}^{n} \left( \left[ g_{x_i}^{\left( 1 \right)1}, g_{x_i}^{\left( 1 \right)2}, \dots, g_{x_i}^{\left( 1 \right)L} \right] \right).
\end{equation}

Alternatively with the convolutional encoder~\citep{peng2020convolutional}, point features are first encoded with a shallow PointNet~\citep{qi2017pointnet} with local max pooling:
\begin{equation}
    g_{x_i}^{\left( 2 \right)} = \max _{x_j\in\mathcal{N} \left ( x_i \right )} h_{\Theta}\left(h_{x_j}, x_j-x_i\right).
\end{equation}
The point-wise features are then projected onto three feature planes and form the latent code $\mathbf{z}^{\left( 2 \right)}$:
\begin{equation}
\label{eq:latent_conv}
\mathbf{z}^{\left( 2 \right)} = \mu_{\Theta} \left( \texttt{project}_u \left( g^{\left( 2 \right)} \right) \right),
\end{equation}
where $u \in \{\text{XY}, \text{XZ}, \text{YZ}\}$ represents the three orthogonal planes. $\mu_{\Theta}$ denotes a U-Net with weights shared across the planes, and $\texttt{project}_u$ indicates projection onto plane $u$.

\subsubsection{Query sampling}

\begin{figure}[htb]
    \centering
    \subfloat[Volume]{\includegraphics[width=0.29\linewidth, keepaspectratio]{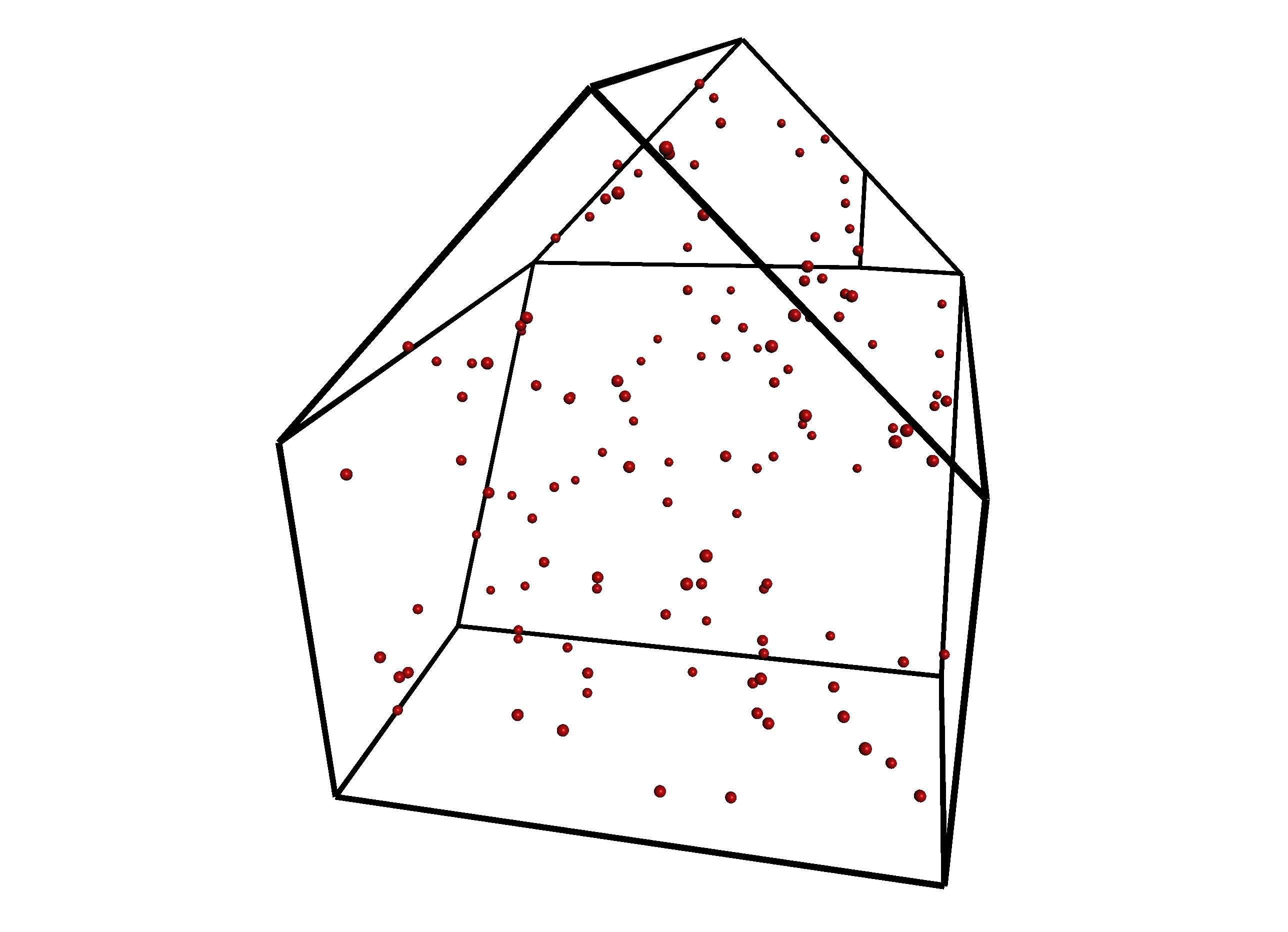}}
    \hspace{.5em}
    \subfloat[Boundary]{\includegraphics[width=0.29\linewidth, keepaspectratio]{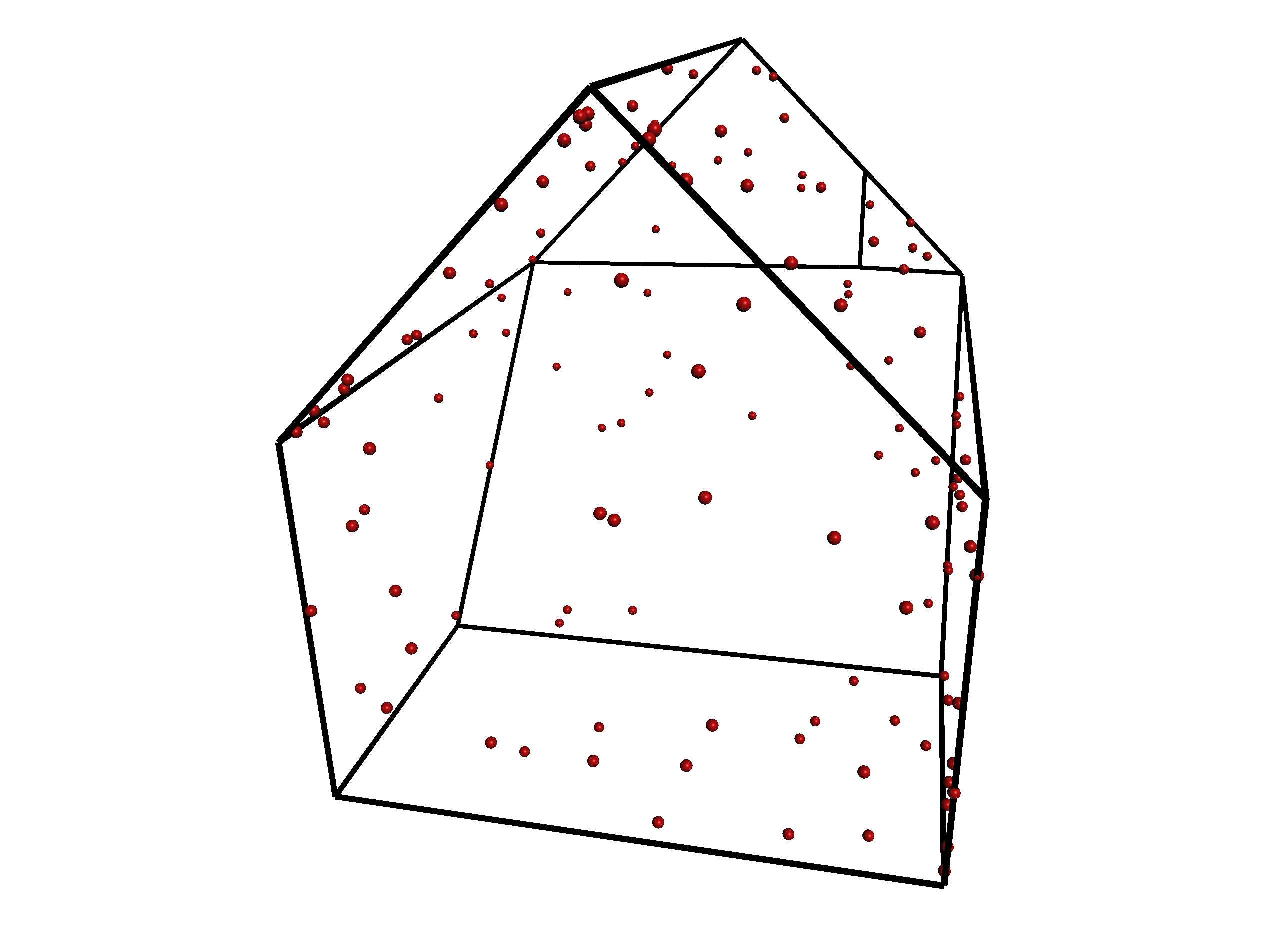}}
    \hspace{.5em}
    \subfloat[Skeleton]{\includegraphics[width=0.29\linewidth, keepaspectratio]{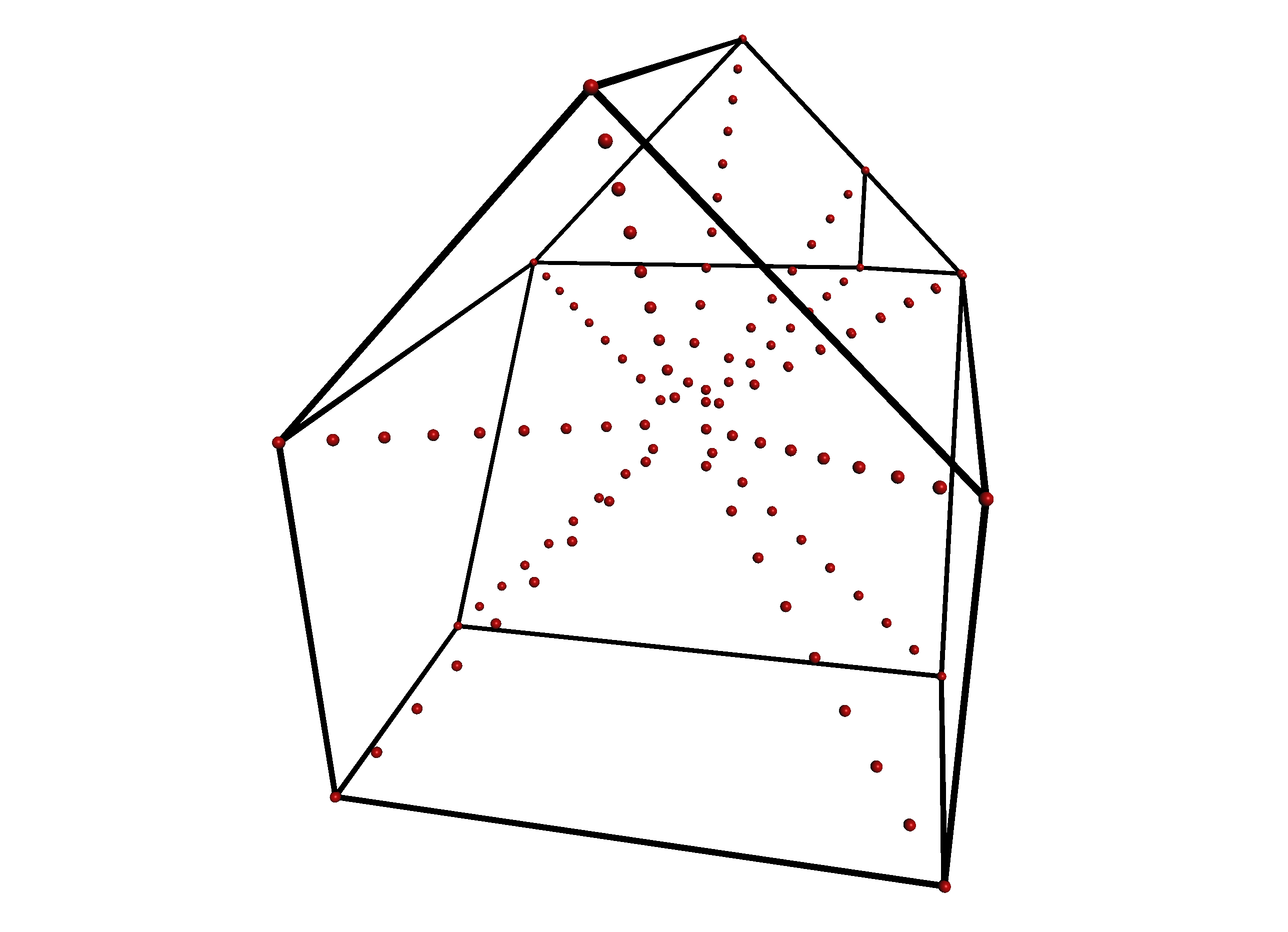}} \\
    \subfloat[Full skeleton representatives]{\includegraphics[width=0.85\linewidth, keepaspectratio]{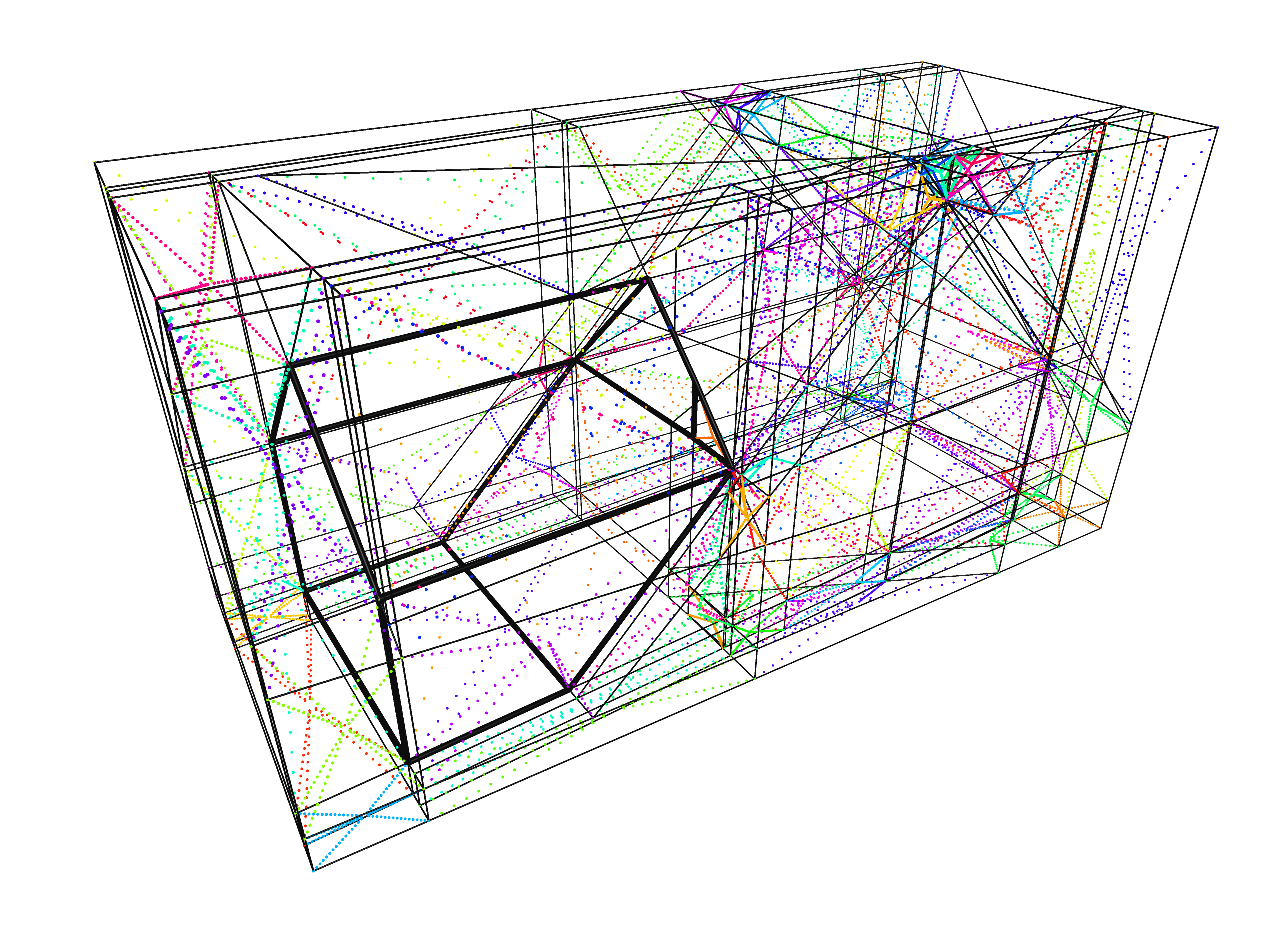}}
    \caption{Sampling representative points from a polyhedron. (a) - (c) visualize different strategies to sample the polyhedron highlighted in (d). \textbf{Volume}: Points are sampled randomly from inside the volume. \textbf{Boundary}: Points are sampled from the boundary. \textbf{Skeleton}: Points are sampled along the polyhedral skeleton as described in ~\autoref{alg:skeleton_sampling}. Representative points are color-coded by their parent polyhedra.}
    \label{fig:sampling}
\end{figure}

To encode an arbitrary-shaped polyhedron, one challenge lies in consistently describing the heterogeneous polyhedral geometry. To address this, we propose sampling representative points from inside the polyhedron and coercing the geometry into fixed-length queries $\mathbf{s}$ of size $k$: $\mathbf{s} = \{s_1, s_2, ..., s_k\}$. It is clear that the more representative the sampled points are, the more information they convey about the polyhedron. We propose skeleton sampling that picks samples from both vertices and principal axes, as described in ~\autoref{alg:skeleton_sampling} and ~\autoref{fig:sampling}. Vertices, because of their prominence in describing sharp geometry, are prioritized over points along the axes when a low value of $k$ is given.

\begin{algorithm}[htb]
    {
	\KwIn{Vertices $\mathcal{V}$, centroid $C$, and \#samples $k$}
	\KwOut{Representative points $\mathbf{s}$}
	\BlankLine

	$\mathbf{s}$ $\leftarrow$ $\operatorname{init} \emptyset$\;

        \eIf{$k \leq |\mathcal{V}|$}
        {
    	$\mathcal{V}_s \leftarrow \operatorname{sample}_k(\mathcal{V})$\;
        $\mathbf{s}$ $\leftarrow$ $\mathcal{V}_s$ \;
        }
        {
        $k_m$ $\leftarrow$ $\lfloor \frac{k}{|\mathcal{V}|} \rfloor$ \;
        $k_l$ $\leftarrow$ $k \bmod |\mathcal{V}|$ \;
        $\mathbf{s}_m$ $\leftarrow$ $ \operatorname{sample}_{k_m}  \{ \left ( \mathcal{V}_i, C \right ): i = 1, 2, ..., |\mathcal{V}|-1\}$ \;
        $\mathbf{s}_l$ $\leftarrow$ $ \operatorname{sample}_{k_l} \left ( \mathcal{V}_{|\mathcal{V}|}, C \right ) $ \;
        $\mathbf{s}$ $\leftarrow$ $\{ \mathbf{s}_m, \mathbf{s}_l \}$ \;
        }
		\Return{$\mathbf{s}$}
		
    }
	\caption[Sample]{S\textsc{keleton sampling} ($\mathcal{V}$, $C$, $k$)}
	\label{alg:skeleton_sampling}
\end{algorithm}

We also evaluate two other sampling strategies, namely volume sampling and boundary sampling, as illustrated in \autoref{fig:sampling}. The volume variant randomly takes points inside the volume of a polyhedron, carrying relatively the least amount of geometric information about the polyhedron. The boundary variant samples points on the boundary of a polyhedron with area-induced probability. This variant can better depict polyhedral occupancy with boundary information. Since the skeleton variant picks samples from both vertices and principal axes, it provides arguably the most efficient description of a polyhedron among the three variants.

The representative points obtained by any of the three sampling strategies reduce the complexity of an arbitrary-shaped polyhedron to a fixed-size feature vector that can be consumed by the neural network while preserving geometric information to different extents. These representative points then serve as queries against $\mathbf{z}$, leading to the formation of polyhedron-wise features. Intuitively, these queries are used to jointly describe the occupancy of the underlying polyhedron. Note that each variant is applied individually, and their performances are compared in \autoref{sec:ablation}.

\subsubsection{Forming polyhedron-wise features}

\begin{figure}[htb]
  \centering
  \centerline{\epsfig{figure=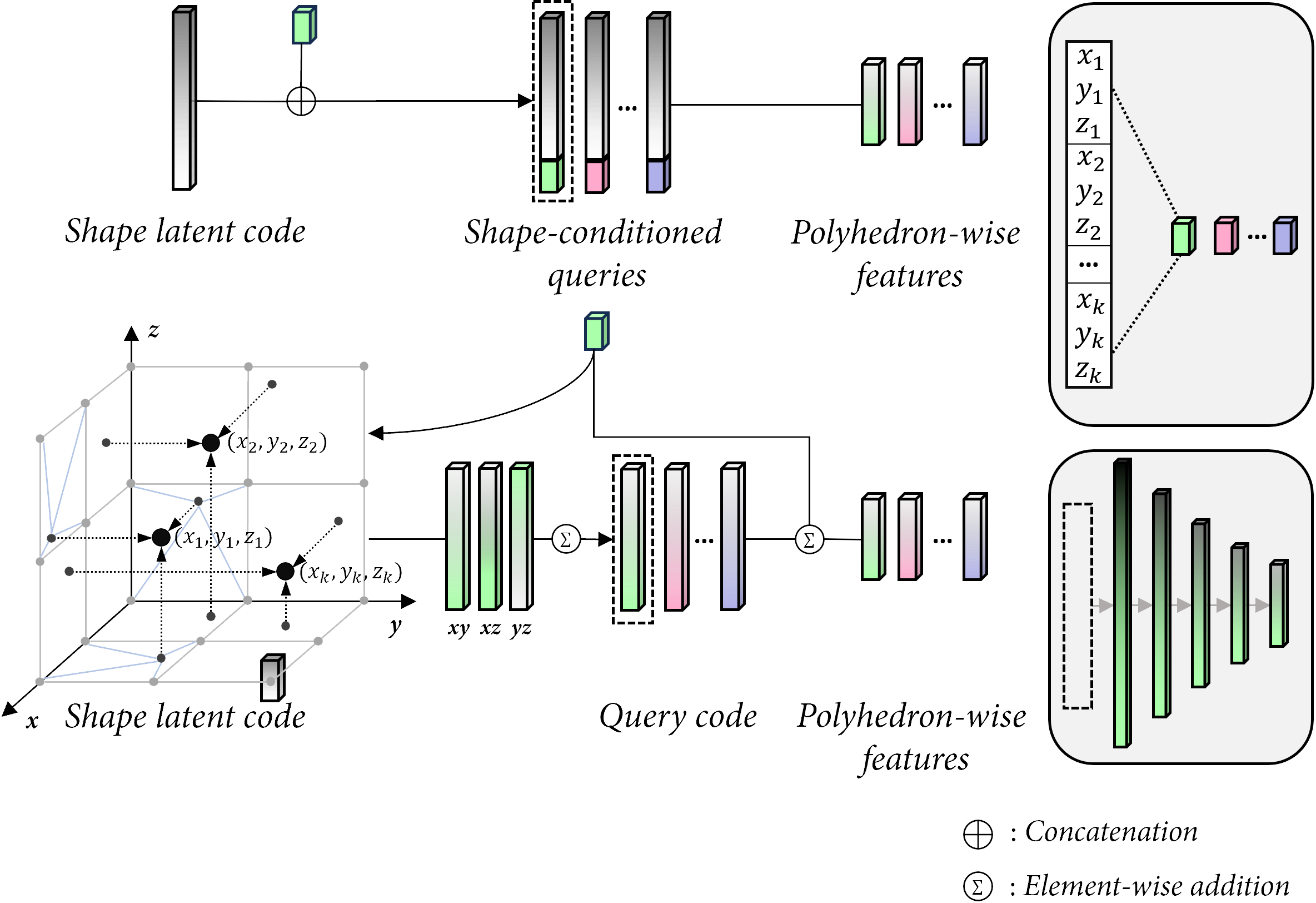,width=0.99\linewidth}}
\caption{Fusion of shape latent code and polyhedral queries to form polyhedron-wise features, with the plain encoder (top) and with the convolutional encoder (bottom).}
\label{fig:fusion}
\end{figure}

Inspired by the recent advance in 3D shape representation learning~\citep{park2019deepsdf,yao20213d}, for the latent code $\mathbf{z}^{\left( 1 \right)}$, we form a shape-conditioned implicit representation $\mathbf{z}^{\left( 1 \right)}_s$ of the polyhedron by concatenating the coordinates of the queries $\mathbf{s}$ with $\mathbf{z}^{\left( 1 \right)}$:
\begin{equation}
\mathbf{z}^{\left( 1 \right)}_\mathbf{s} = \xi_{\Theta} \left(\mathbf{s} \mid \mathbf{z}^{\left( 1 \right)} \right) = \xi_{\Theta} \left( \left[s_1, s_2, ..., s_k, \mathbf{z}^{\left( 1 \right)} \right] \right),
\label{eq:feature_plain}
\end{equation}
where $\xi_{\Theta}$ represents an MLP. 

Alternatively, the conditioned implicit representation $\mathbf{z}^{\left( 2 \right)}_s$ of the latent code $\mathbf{z}^{\left( 2 \right)}$ is constructed by interpolating $\mathbf{z}^{\left( 2 \right)}$ at the coordinates of $\mathbf{s}$:
\begin{equation}
\mathbf{z}^{\left( 2 \right)}_\mathbf{s} = \xi_{\Theta} \left( \mathbf{s} \mid \mathbf{z}^{\left( 2 \right)} \right) = \xi_{\Theta} \sum_{u} \left( \texttt{interpolate}_{\mathbf{z}^{\left( 2 \right)}} \left( \mathbf{s} \right) \right),
\label{eq:feature_conv}
\end{equation}
where $\texttt{interpolate}$ denotes bilinear interpolation, and $\mathbf{s}$ represents the coordinates of queries. Both \autoref{eq:feature_plain} and \autoref{eq:feature_conv} enable the modeling of multiple building instances with a single neural network. \autoref{fig:fusion} illustrates the different formations of polyhedron-wise features with $\mathbf{z}^{\left( 1 \right)}_s$ and $\mathbf{z}^{\left( 2 \right)}_s$.

Intuitively, $\mathbf{z}_s$ represents a discrete occupancy function that, given a polyhedron, describes its occupancy conditioned on the underlying building instance. This representation can be interpreted as a spatial classifier for which the decision boundary is the surface of the building. Notably, instead of approximating a continuous implicit function by exhaustive enumeration, our discretized formulation takes geometric priors of individual polyhedra into account, which significantly reduces computational complexity and mitigates solution ambiguity. \autoref{fig:difference} illustrates this distinction.

\subsection{Graph node classification}

The polyhedron-wise features produced by \autoref{eq:feature_plain} or \autoref{eq:feature_conv} do not yet account for inter-polyhedron adjacency, which could provide additional information for classifying individual polyhedra. To leverage this topological information and enhance classification performance, we utilize another stack of graph convolution layers for graph node classification, as outlined in \autoref{eq:classification}.
Specifically, we employ topology-adaptive graph convolution \citep{du2017topology} for its adaptivity to the topology of the graph and computational efficiency. It utilizes a set of fixed-size learnable filters for graph convolution, defined as follows:
\begin{equation}
\mathbf{G}^{l+1}=\sum_{k=0}^K\left(\mathbf{D}^{-1 / 2} \mathbf{A D}^{-1 / 2}\right)^k \mathbf{G}^{l},
\label{eq:convolution}
\end{equation}
where $\mathbf{G}^{l+1}$ and $\mathbf{G}^{l}$ denote the node features before and after the convolution at the $l$-th layer, respectively. $\mathbf{A}$ is the adjacency matrix implied by $\mathcal{E}$ in \autoref{eq:classification}, and $\mathbf{D}=\operatorname{diag}[\mathbf{d}]$ with the $i$-th component being $d\left(i\right) = \sum_j \mathbf{A}_{i, j}$. $K$ is the number of filters, whose topologies are adaptive to the topology of the graph.

Multiple graph convolution layers defined in \autoref{eq:convolution} are stacked to increase the receptive field of the neural network. The feature of the $i$-th node $G_i$ is then fed into a binary classification head with the softmax activation function to produce the probability of the polyhedron $v_i$ being \emph{interior}:
\begin{equation}
    \hat{y}_i = \operatorname{softmax} \left( \lambda_{\Theta} \left( G_i \right) \right),
\label{eq:prediction}
\end{equation}
where $\lambda_{\Theta}$ denotes an MLP.

\begin{figure}[htb]
  \centering
  \centerline{\epsfig{figure=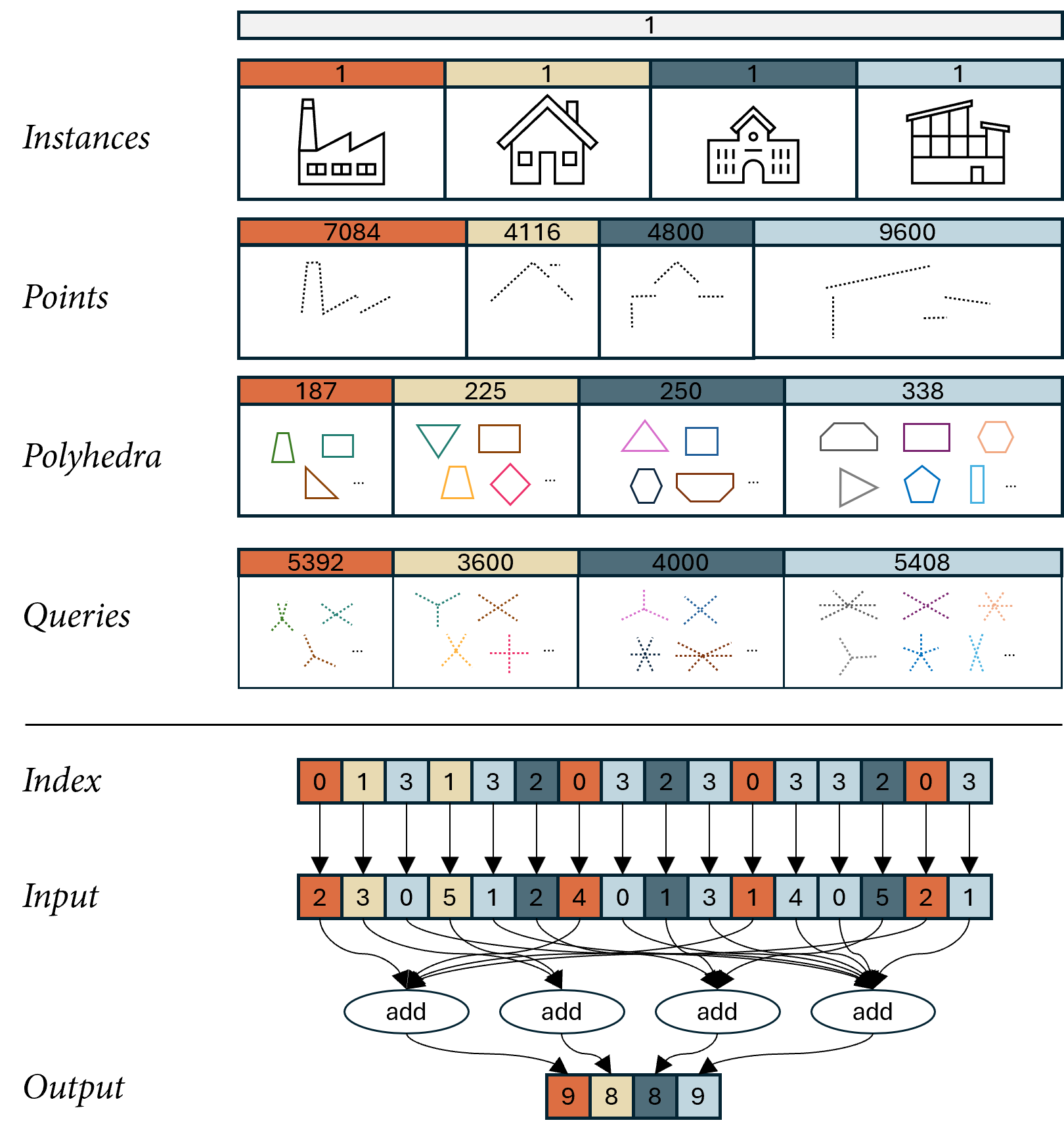,width=0.99\linewidth}}
\caption{PolyGNN efficiently accommodates variable-size input points, polyhedra, and queries with an index-driven batching technique (top). Batch size equals 4 in this example. An example with scattered \texttt{add} operator (bottom).}
\label{fig:batch}
\end{figure}

The number of input points, polyhedra, and queries may vary considerably among different building instances,  naturally impeding parallelization with mini-batches. To mitigate this variability, we employ a batching mechanism tailored to accommodate such diversity. Our approach entails recording indices of the points, polyhedra, and queries within each batch. These indices are subsequently utilized to acquire instance-level features, as depicted in \autoref{fig:batch}.

The network can be supervised by cross entropy or focal loss~\citep{lin2017focal} that minimizes the discrepancy between the prediction and the ground truth. It can be optimized end-to-end without any auxiliary supervision.
In the testing phase, given a building instance, we predict the occupancy of the candidate polyhedra. Then the surface lies in between pairs of polyhedra $\{v_i, v_j\}$ with different class predictions, i.e., $y_i \neq y_j$, as shown in \autoref{fig:partition}.

\section{Experimental settings}
\label{sec:settings}

\subsection{Datasets}

\begin{figure}[t]
  \centering
  \centerline{\epsfig{figure=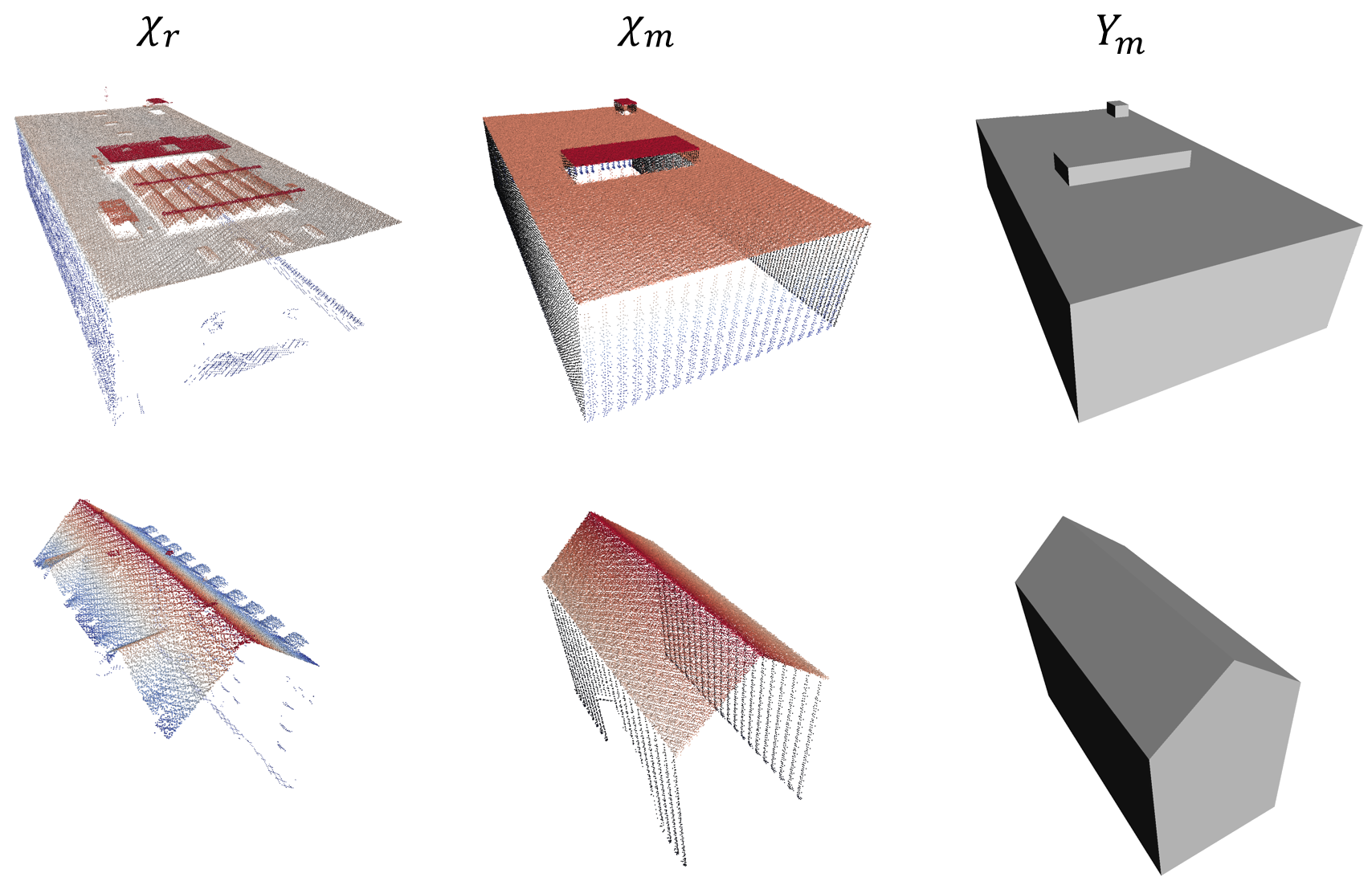,width=0.9\linewidth}}
\caption{Examples of abstraction gaps between real-world point clouds $\mathcal{X}_r$ and existing building models $Y_m$. Instead of learning $f$: $\mathcal{X}_r \rightarrow Y_m$, we learn an auxiliary mapping $f^\prime$: $\mathcal{X}_m \rightarrow Y_m$, where $\mathcal{X}_m$ is derived from $Y_m$ by synthesizing $\mathcal{X}_r$. Point clouds are color-coded by their height fields.}
\label{fig:abstraction}
\end{figure}

Unlike other applications that adhere to rigorous definitions of ground truths, reconstructing large-scale polygonal buildings presents a challenge due to the abstraction of existing building models~\citep{wang2023building3d,wichmann2018roofn3d}, resulting in inevitable deviations from actual measurements, as shown in \autoref{fig:abstraction}. This abstraction would impede a supervised learning algorithm due to its inherent biases. To overcome this abstraction gap, we created a synthetic dataset comprised of simulated airborne LiDAR point clouds and their corresponding building mesh models. This dataset enables a reliable mapping between the two sources, enabling a fair evaluation of the proposed method for primitive assembly.

Formally, let $\mathcal{X}_r$ and $Y_m$ be a real-world point cloud and its corresponding building model, respectively. Due to the abstraction gap, the mapping $f$: $\mathcal{X}_r \rightarrow Y_m$ cannot be accurately learned by a neural network. Instead, we opt to learn an auxiliary mapping $f^\prime$: $\mathcal{X}_m \rightarrow Y_m$ where $\mathcal{X}_m$ is derived from $Y_m$ by synthesizing $\mathcal{X}_r$. Once $f^\prime$ is learned, it can be applied to $\mathcal{X}_r$ to obtain the corresponding output $Y_r^\prime = f^\prime(\mathcal{X}_r)$. Conditioned on good transferability to real-world point clouds, using synthetic data in our task offers two-fold advantages. First, it enables the learning of the desired mapping by circumventing the abstraction, allowing the classifier to be trained and evaluated independently of potential data discrepancies. Moreover, it facilitates the exploration of a large volume of ``free'' training data, which benefits the learning algorithm in general.

We utilize the \textit{Helios++} simulation toolkit \citep{winiwarter2022virtual} to simulate airborne LiDAR scanning. LoD2 building models from Bavaria, Germany are used as references for their high quality and coverage \citep{bavaria2022}. Artifacts such as noise and inter-building occlusions are intentionally included in the scanning process, to assimilate the distribution of $\mathcal{X}_m$ and $\mathcal{X}_r$, thereby enhancing the robustness of the neural network against real-world measurements. The virtual sensor closely emulates the characteristics of \textit{Leica HYPERION2+}, utilizing an oscillating optics system with a pulse frequency of 1.5\,MHz and a scan frequency of 150\,Hz. We simulate an airborne survey performed by a \textit{Cirrus SR22} aircraft flying at an altitude of 400\,m with a strip interval of 160\,m.
Our training dataset comprises 281,571 buildings with a total of 6,532,880,764 points extracted from the city of Munich, Germany, with an additional 10,000 buildings reserved for evaluation. On average, each building in the dataset is associated with 22,406 points. To assess the cross-city transferability of PolyGNN, we also synthesize data from 220,127 buildings in Nuremberg, Germany. In addition to the synthetic data, we apply the trained model directly to a real-world airborne LiDAR point cloud dataset containing 1,452 buildings captured with \textit{Leica HYPERION2+}. For an individual building, we normalize it to unit scale, generate a set of polyhedra with inter-polyhedron adjacency as pre-processing (see ~\autoref{subsec:graph}), and use ray tracing to determine the ground truth occupancy label for every polyhedron.

\subsection{Evaluation metrics}

We utilize multiple criteria to evaluate the performance of the reconstruction. The classification accuracy directly impacts the fidelity of the reconstruction and is therefore evaluated. Furthermore, since the ground truths are reliably defined in our setting, we quantify the surface discrepancy between the reconstructed surface and the ground truth by calculating the Hausdorff distance $H$:
\begin{equation}
    H = \max \left\{ \sup_{a \in A} \inf_{b \in B} d(a, b), \sup_{b \in B} \inf_{a \in A} d(a, b) \right\},
\label{eq:hausdorff}
\end{equation}
where $d(a, b)$ represents the distance between points $a$ and $b$. We randomly sample 10,000 points from both the reconstructed surface and the ground truth and calculate both the absolute and relative distances.
We quantify the success rate $S$ by the proportion of 10,000 samples that are solvable. Typical unsolvable cases include empty reconstruction due to the absence of \emph{interior} polyhedra, or timeout. For a fair comparison in the context of large-scale reconstruction, in the event of an unsolvable reconstruction, we assign the length of the largest side of the bounding box as the absolute distance, and 100\% as the relative distance. We also employ RMSE for evaluating fidelity of the reconstruction on real-world data. Additionally, we measure the geometric complexity of the reconstructed building models in terms of the number of faces they comprise, denoted as $N_F$, and measure computational efficiency in terms of running time $t$ with a 5-minute timeout for an individual building.

\subsection{Implementation details}

We implemented adaptive space partitioning with robust Boolean spatial operations from \textit{SageMath} \citep{sagemath}. For query sampling, while a larger value of $k$, representing the number of samples per polyhedron, could enhance the polyhedral representation, particularly for polyhedra with a large number of faces, this increased representativeness would come with additional computational costs. For all of our experiments, we set $k$ to 16 to balance the representativeness and computational complexity.

Although our implementation with the index-driven batching accommodates variable-length input point clouds, unless otherwise specified, the input point clouds are downsampled to 4,096 points. Point clouds are normalized before being fed into the network and rescaled for computing the Hausdorff distance.
All experiments are optimized by the Adam with a base learning rate $10^{-3}$ and weight decay $10^{-6}$, with batch size 64. The network variants were trained for 50 epochs for the ablation experiments, whereas they continued training until 150 epochs for the best model in other experiments.

\section{Results and analysis}
\label{sec:results}

\subsection{Alternative and ablation experiments}
\label{sec:ablation}

As shown in \autoref{tab:polyhedron_sampling}, among the three sampling strategies presented in \autoref{fig:sampling}, skeleton sampling achieves the best classification and geometric accuracy, followed by boundary sampling. This finding aligns with the fact that both skeleton sampling and boundary sampling leverage more explicit geometric information compared to the volume counterpart, with the skeleton of a polyhedron capturing the most critical information conveyed by its vertices and principal axes.

\begin{table}[t]
\centering
\begin{tabular}{ccc}
\hline
Query sampling & Accuracy (\%) $\uparrow$ & H (m) $\downarrow$ \\ \hline
Random         & 94.5        & 1.20    \\
Boundary       & 94.7        & 1.12    \\
Skeleton       & \textbf{95.5}        & \textbf{1.08}    \\ \hline
\end{tabular}
\caption{Impact of query sampling strategy on model performance.}
\label{tab:polyhedron_sampling}
\end{table}

\begin{figure}[t]
  \centering
  \centerline{\epsfig{figure=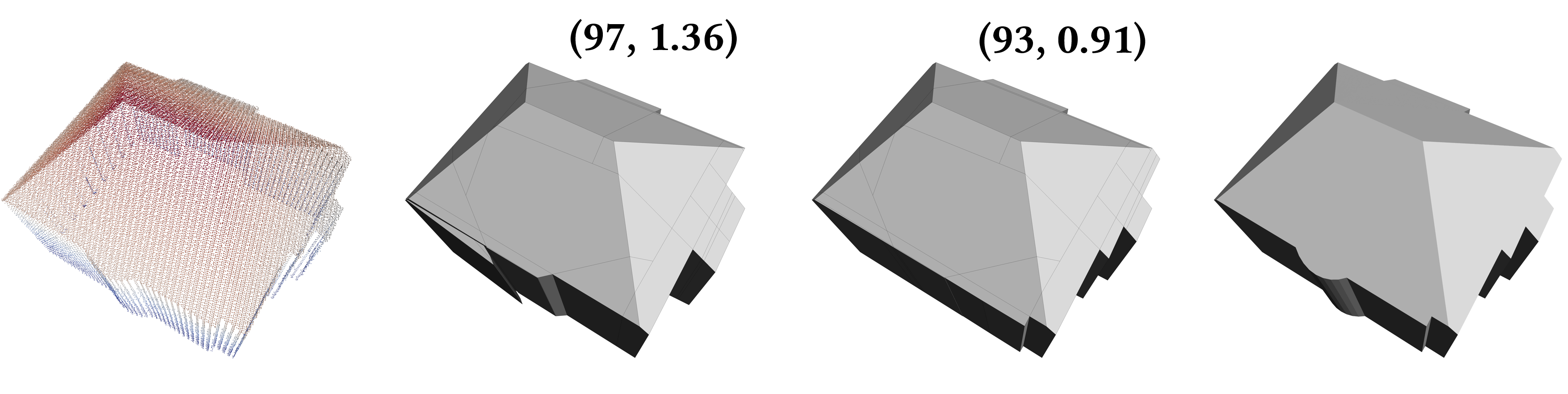,width=0.97\linewidth}}
\caption{Impact of adjacency in PolyGNN reconstruction. From left to right: input point cloud color-coded by height field, reconstructed model w/o adjacency, reconstructed model w/ adjacency, ground truth. $\left( \bullet, \bullet \right)$ denotes $\left( N_F, H \right)$.}
\label{fig:ablation}
\end{figure}

The individual contributions of the classification head and the adjacency information to the reconstruction performance are analyzed through another ablation experiment, as presented in \autoref{tab:adjacency}. Replacing the classification head with a regression head and the use of the $L_2$ loss cause the network to collapse completely, leading to the prediction of every polyhedron as an exterior one. In this case, the classification accuracy represents the high dominance of exterior polyhedra (87.3\%). Furthermore, the results provide clear evidence that incorporating inter-polyhedron adjacency information significantly enhances the reconstruction performance compared to relying solely on monotonic polyhedral information (95.5\% vs. 93.7\%). This improvement suggests that PolyGNN effectively exploits neighborhood information for occupancy estimation. Additionally, \autoref{fig:ablation} visually demonstrates the effectiveness of such information where the network utilizes adjacency information to achieve a more regularized reconstruction. This regularization is analogous to the MRF employed in \citet{chen2022points2poly}, while with PolyGNN it is integrated into the feature space, avoiding additional computational overhead.

\begin{table}[t]
\centering
\begin{tabular}{cccc}
\hline
Classification        & Adjacency             & Accuracy (\%) $\uparrow$ & H (m) $\downarrow$ \\ \hline
\xmark                & \xmark                &  87.3    &    -     \\
\cmark                & \xmark                &  93.7    &  1.80    \\
\cmark                & \cmark                &  \textbf{95.5}  &  \textbf{1.08}    \\ \hline
\end{tabular}
\caption{Impact of classification head and adjacency information on model performance. ``-'' indicates complete failure where no model is reconstructed.}
\label{tab:adjacency}
\end{table}

PolyGNN is designed to be agnostic to the number of points, allowing for point clouds with varying sizes as inputs with advanced mini-batching. In \autoref{tab:point_sampling}, we compare three point cloud sampling options: random sampling, coarse grid sampling with a resolution of 0.05 within a unit cube, and fine grid sampling with a resolution of 0.01 within the same unit cube. The results demonstrate that random sampling outperforms grid sampling with both resolutions in terms of accuracy. It is worth noting that our random sampling strategy is dynamic, where different random points are selected in different epochs during training. This dynamic random sampling can also be considered a form of data augmentation. Although computationally more efficient, coarse grid sampling does not entail sufficient details for input point clouds. Interestingly, fine grid sampling leads to significantly longer training times, yet it yields lower accuracy compared to random sampling, possibly due to the inherent difficulty of encoding shape latent codes from variable-length inputs.

\begin{table}[t]
\centering
\begin{tabular}{ccccc}
\hline
Point sampling     &  Accuracy (\%) $\uparrow$ & Time \textit{train} (h.) $\downarrow$ \\ \hline
Grid (res. 0.05)   & 94.6                      & \textbf{0.7}     \\
Grid (res. 0.01)   & 94.8                      & 24.6    \\
Random             & \textbf{95.5}             & 2.1     \\  \hline
\end{tabular}
\caption{Impact of point cloud sampling strategy on model performance and per-epoch training time. ``res'' represents grid resolution relative to a unit cube.}
\label{tab:point_sampling}
\end{table}

Table \ref{tab:backbone} compares the performance of the two encoders, demonstrating the superiority of the convolutional encoder over the plain encoder in terms of both accuracy and efficiency. This advantage reveals that the latent code generated by \autoref{eq:latent_conv} preserves more local information compared to the shape latent code described in \autoref{eq:latent_plain}. Unless otherwise stated, the subsequent experimental analysis utilized the convolutional encoder.

\begin{table}[t]
\centering
\begin{tabular}{ccccc}
\hline
Encoder       &  Accuracy (\%) $\uparrow$ & Time \textit{train} (h.) $\downarrow$ \\ \hline
Plain          & 95.5                      & 2.1                                  \\
Convolutional  & \textbf{96.5}             & \textbf{0.4}                         \\ \hline
\end{tabular}
\caption{Impact of encoder on model performance and per-epoch training time.}
\label{tab:backbone}
\end{table}

\subsection{Performance and transferability}

\begin{figure*}[h!]
  \centering
  \centerline{\epsfig{figure=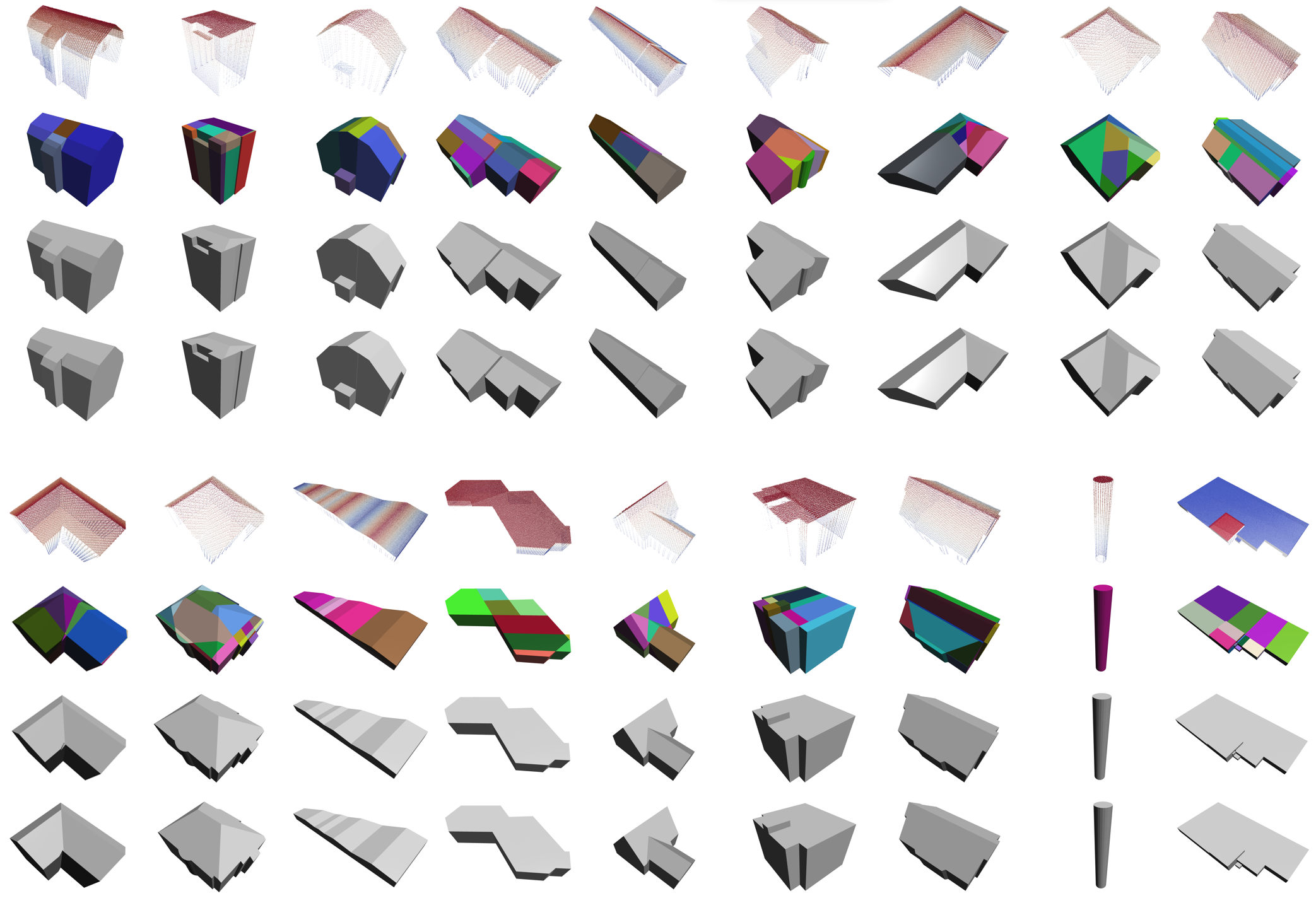,width=16cm}}
\caption{Reconstruction examples on the Munich data with PolyGNN. From top to bottom: input point cloud color-coded by height field, polyhedra classified as building components, reconstructed model, and ground truth model. Point clouds are rendered by their height fields. Polyhedra are randomly color-coded.}
\label{fig:results_munich}
\end{figure*}

PolyGNN achieves an average error of 0.33\,m on the held-out Munich evaluation set. Notably, we observe that when building instances demonstrate similar levels of geometric complexity, accurate classification often leads to lower geometric errors. The reconstructed building models, as shown in \autoref{fig:results_munich}, demonstrate conformity to the distribution of the point clouds while maintaining compactness for potential downstream applications. Buildings with simpler geometry are of more regularity in the reconstruction.

\begin{figure*}[h!]
  \centering
  \centerline{\epsfig{figure=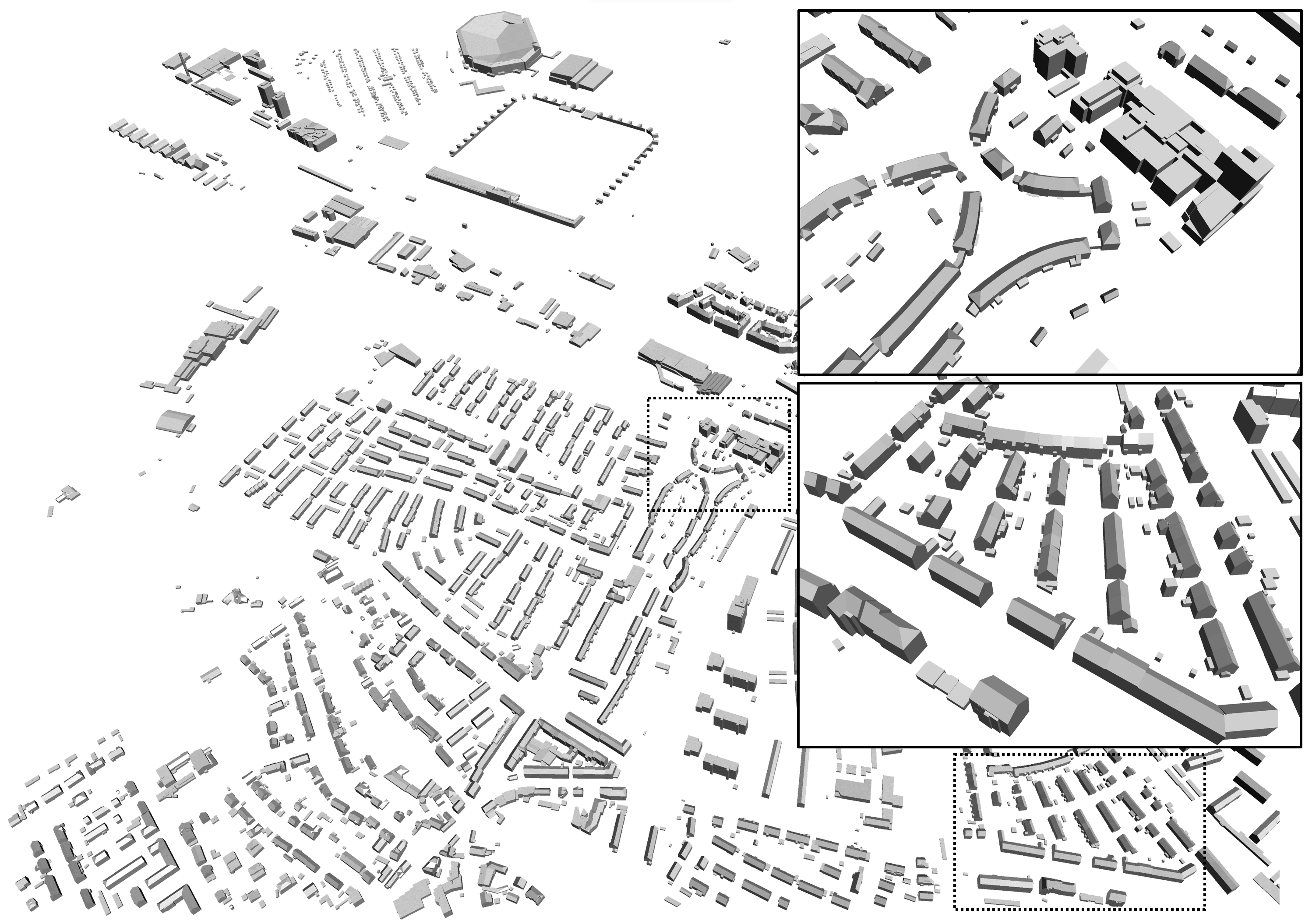,width=0.95\linewidth}}
\caption{Reconstruction of Nuremberg downtown buildings with PolyGNN.}
\label{fig:results_nuremberg}
\end{figure*}

To assess the transferability of PolyGNN, we applied the model trained on the Munich data to buildings in Nuremberg. \autoref{fig:results_nuremberg} showcases the reconstruction of a downtown area of Nuremberg. The Hausdorff distance measures 0.40\,m. The comparable accuracy demonstrates the strong cross-city transferability of our approach when confronted with buildings that may vary in architectural styles. The inference with the convolutional encoder takes 15 seconds for 4,185 buildings in the area, highlighting its efficiency for large-scale reconstruction.

\begin{figure*}[h!]
  \centering
  \centerline{\epsfig{figure=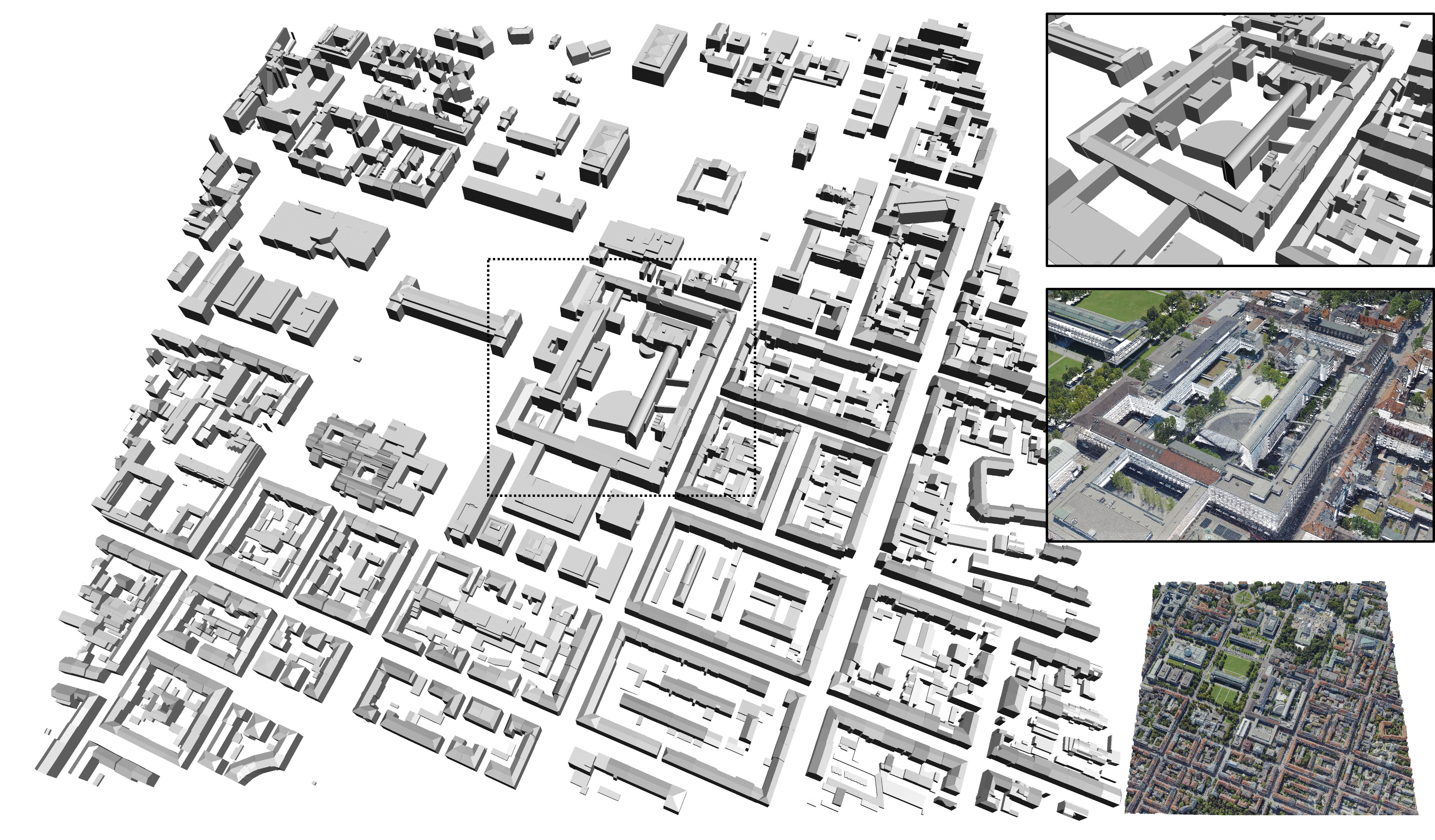,width=0.95\linewidth}}
\caption{Reconstruction from real-world point clouds with PolyGNN trained on the synthetic dataset.}
\label{fig:results_campus}
\end{figure*}

\begin{figure}[h!]
  \centering
  \centerline{\epsfig{figure=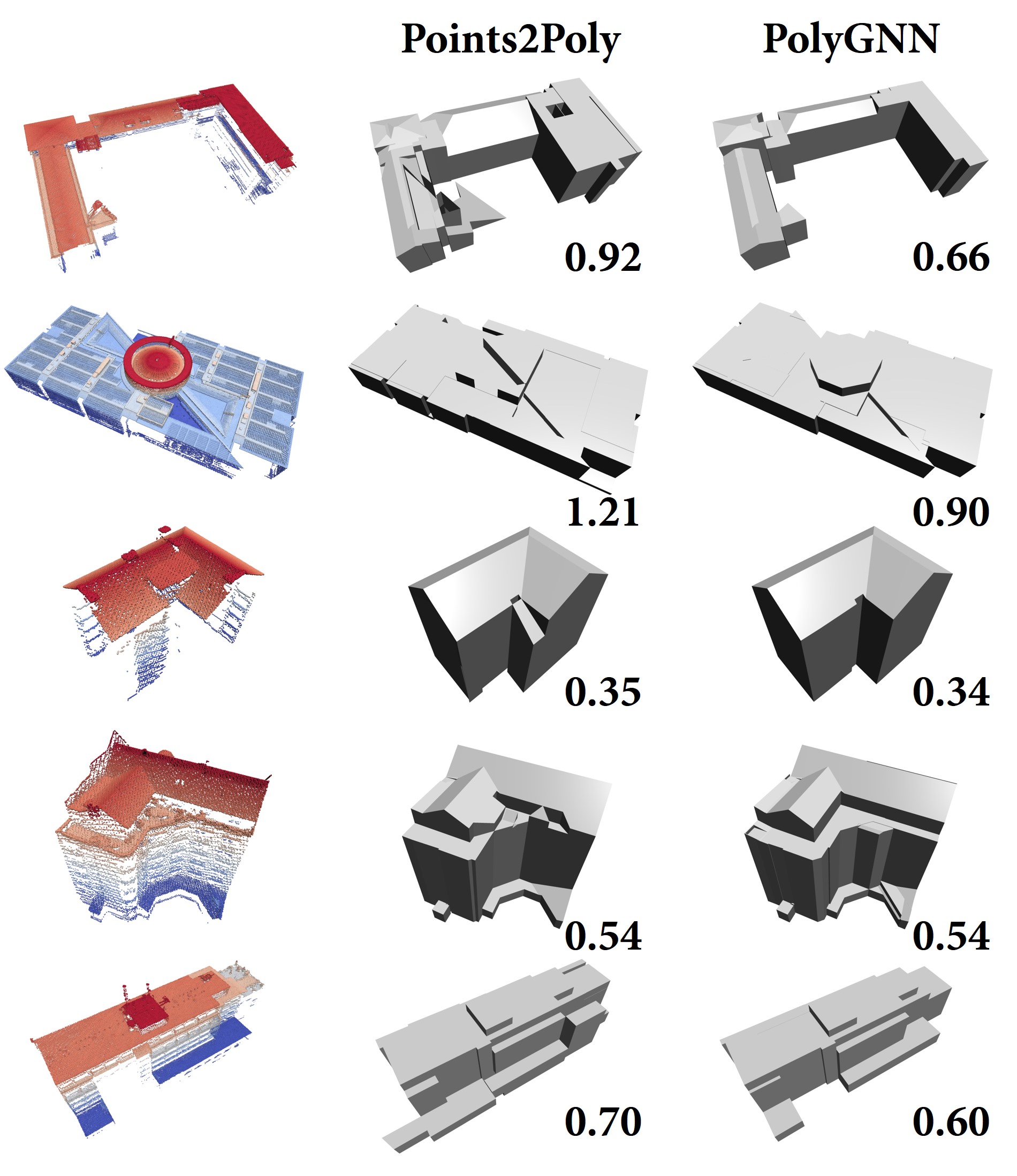,width=0.80\linewidth}}
\caption{Reconstruction from real-world point clouds. Numbers represent RMSE. Both Points2Poly~\citep{chen2022points2poly} and PolyGNN (ours) were trained only on the synthetic data. The models produced by PolyGNN demonstrate lower RMSE.}
\label{fig:transferability}
\end{figure}

\begin{figure}[htb]
  \centering
  \centerline{\epsfig{figure=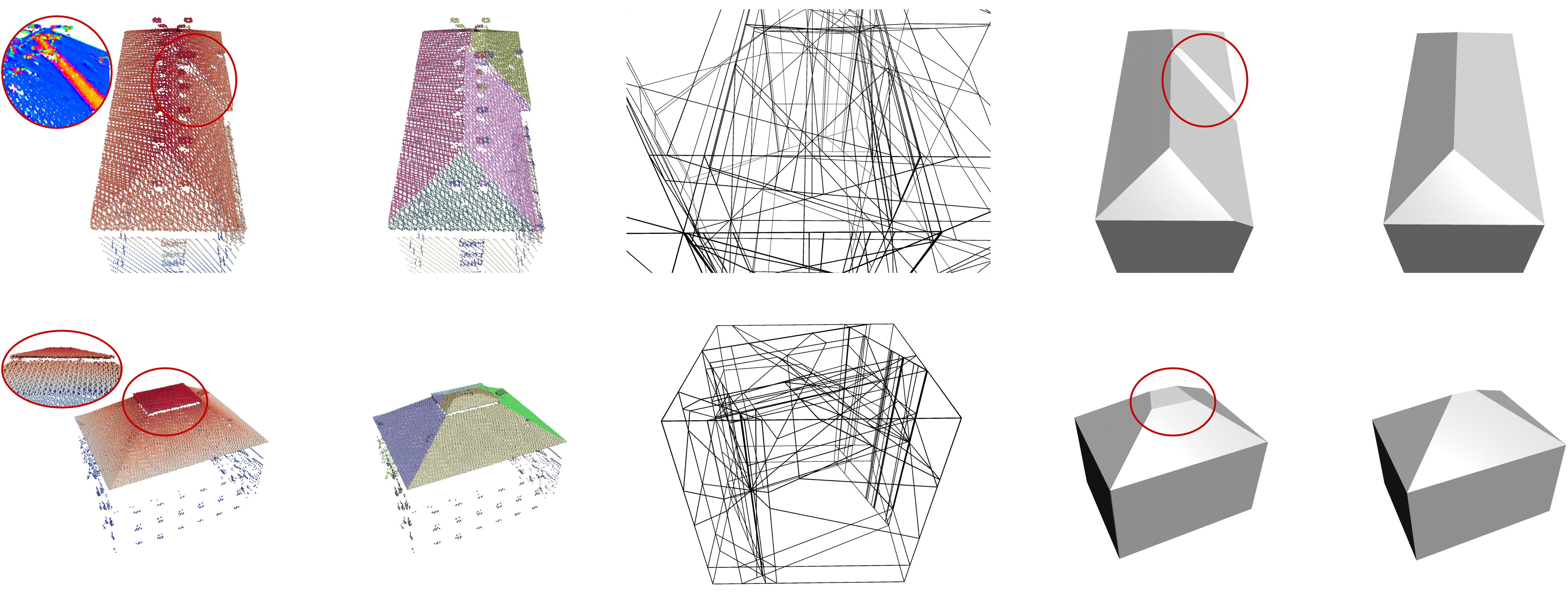,width=0.99\linewidth}}
\caption{Reconstruction from real-world point clouds with planar primitives extracted by RANSAC. From left to right: input point cloud color-coded by height field, the same input color-coded by primitives, cell complex, reconstructed model, and the ground truth. As revealed by the close-up views on the input point cloud, the reconstruction recovered more detailed structures than the ground truth.}
\label{fig:ransac}
\end{figure}

\autoref{fig:results_campus} depicts the reconstruction results obtained by further applying PolyGNN trained exclusively on the synthetic data to real-world point clouds in Munich. \autoref{fig:transferability} shows detailed examples. As expected, a domain gap exists between the two datasets, resulting in suboptimal reconstructions for certain buildings, especially those with architectural styles that are less represented in the training data. Nevertheless, it is noteworthy that the majority of the reconstructed buildings align well with the distribution of the input point clouds. \autoref{fig:ransac} further demonstrates cases where we apply the trained model with extracted planar primitives by RANSAC~\citep{schnabel2007efficient}. By learning the underlying mapping, the reconstruction may approximate the point cloud distribution closer than the ground truth does, which validates the effectiveness of our strategy of learning the auxiliary mapping.

\subsection{Comparison with state-of-the-art methods}

\begin{figure*}[h!]
  \centering
  \centerline{\epsfig{figure=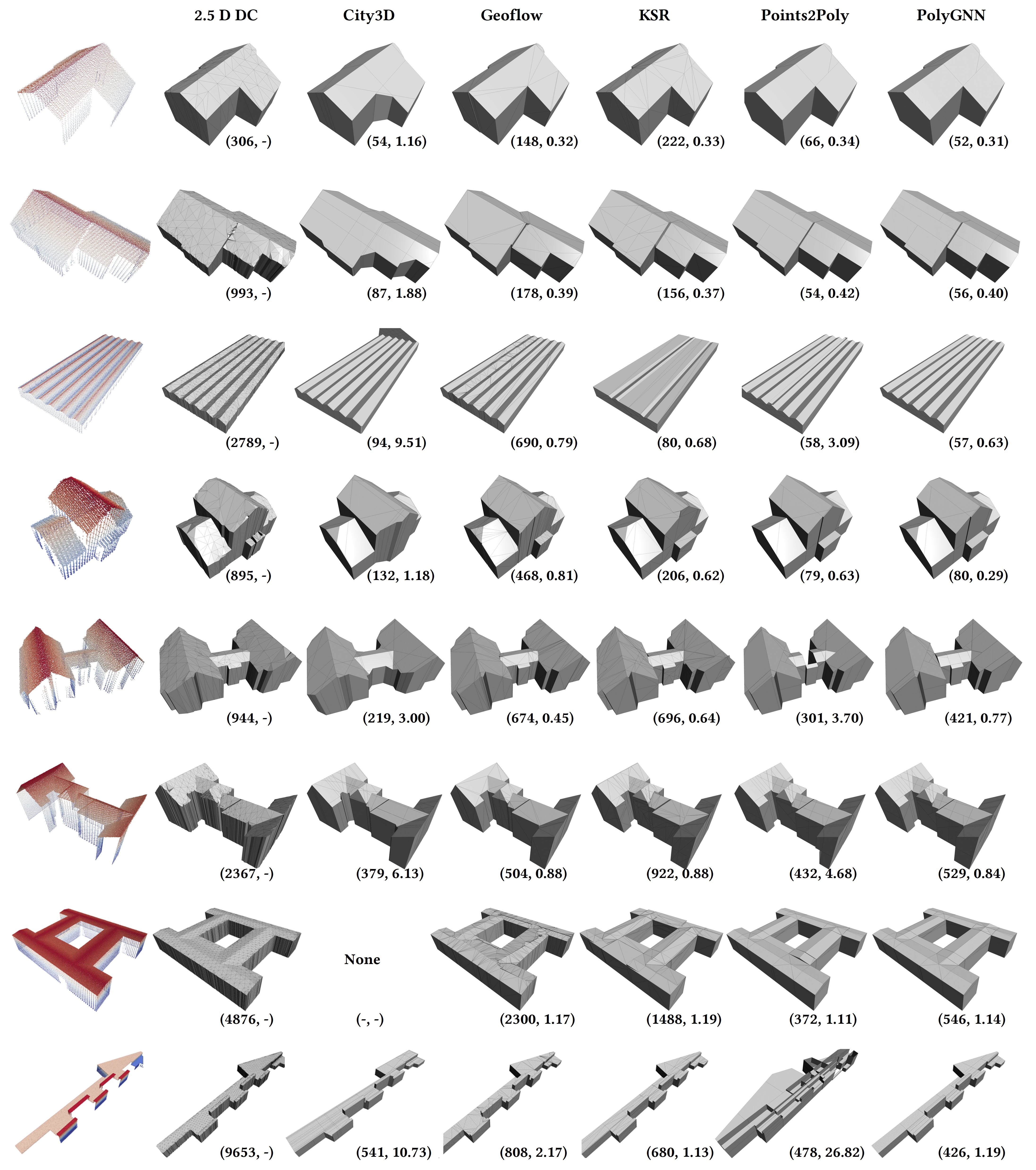,width=0.95\linewidth}}
\caption{Qualitative performance comparison with state-of-the-art methods: 2.5D DC~\citep{zhou20102}, City3D~\citep{huang2022city3d}, Geoflow~\citep{peters2022automated}, KSR~\citep{bauchet2020kinetic}, Points2Poly~\citep{chen2022points2poly}, and PolyGNN (ours). Point clouds are color-coded by their height fields. $\left( \bullet, \bullet \right)$ denotes $\left( N_F, H \right)$. Note that additional wall points were required by KSR; otherwise, it would generates trivial results (see \autoref{fig:comparison_ksr}).}
\label{fig:comparison}
\end{figure*}

\begin{figure}[htb]
  \centering
  \centerline{\epsfig{figure=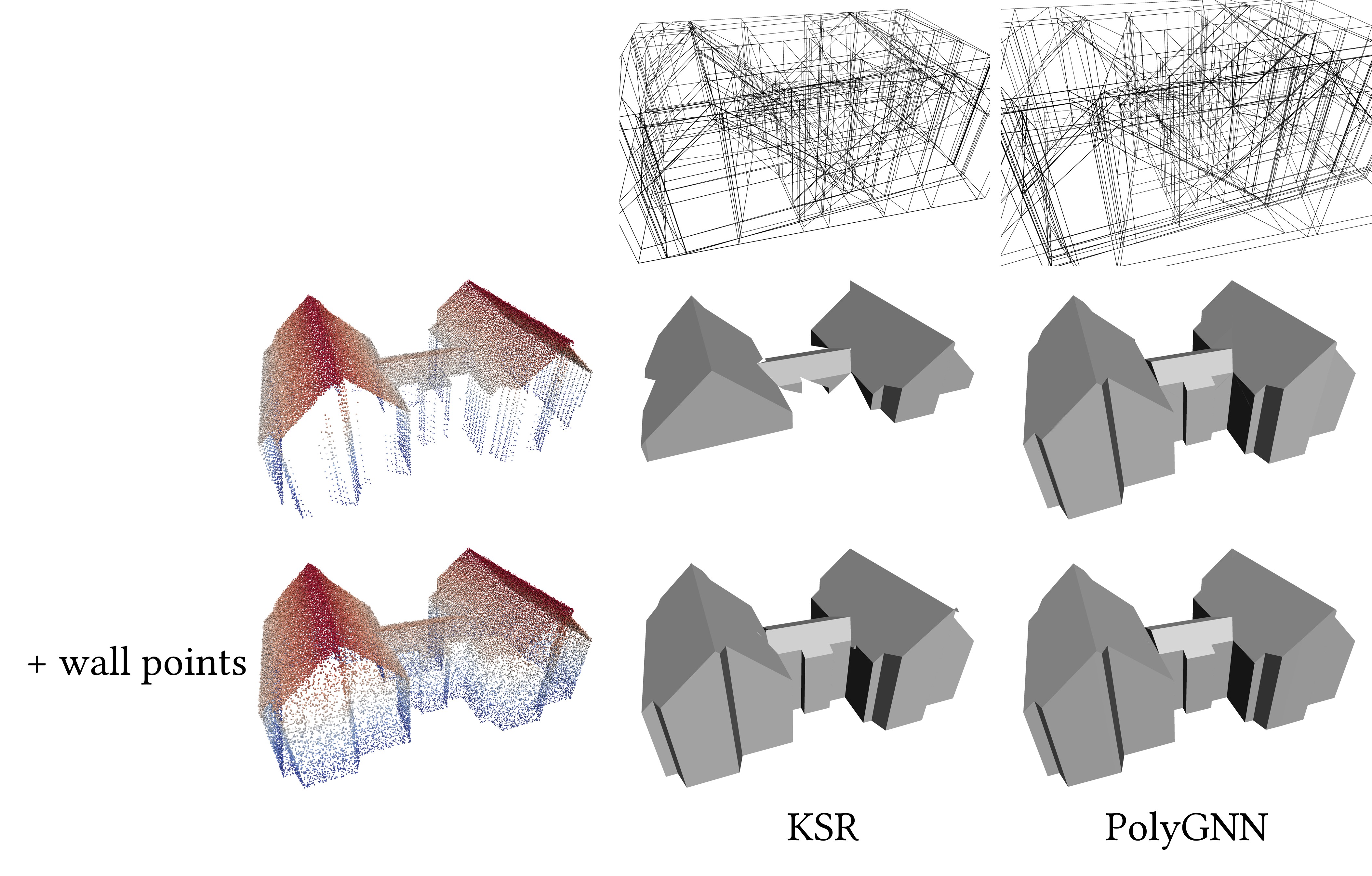,width=0.99\linewidth}}
\caption{Comparison with KSR~\citep{bauchet2020kinetic}. Additional wall points were required by KSR for non-trivial results, in addition to point normals. Given the same planar primitives, PolyGNN (ours) can estimate building occupancy from unorganized incomplete points, and can still deal with added wall points even though it was only trained on airborne data without wall points. Point clouds are color-coded by their height fields.}
\label{fig:comparison_ksr}
\end{figure}

\begin{figure}[htb]
  \centering
  \centerline{\epsfig{figure=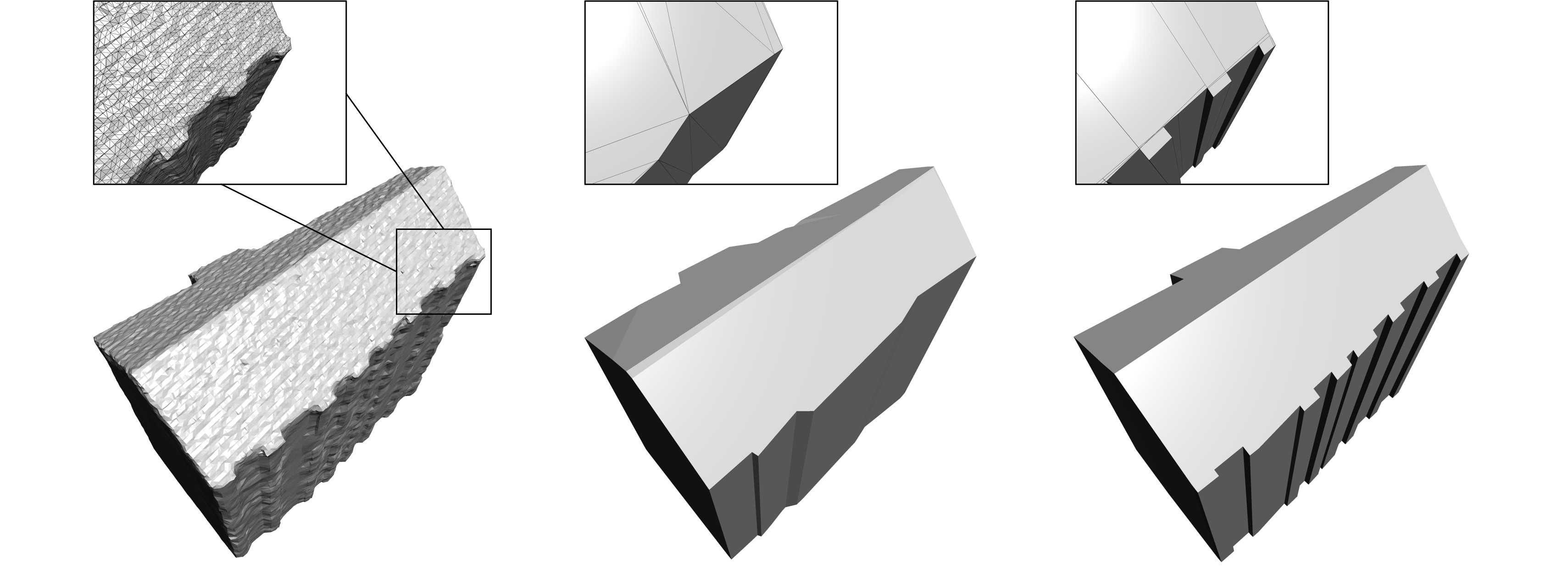,width=0.90\linewidth}}
\caption{Comparison with geometric simplification. Left: smooth surface mesh with Screened Poisson~\citep{kazhdan2013screened} ($N_F = 40042$). Middle: simplified mesh with LowPolyBuildings~\citep{gao2022low} ($N_F = 187$). Right: direct reconstruction with PolyGNN (ours) ($N_F = 190$). Our results demonstrate greater regularity.}
\label{fig:comparison_simplification}
\end{figure}

\begin{figure}[htb]
  \centering
  \centerline{\epsfig{figure=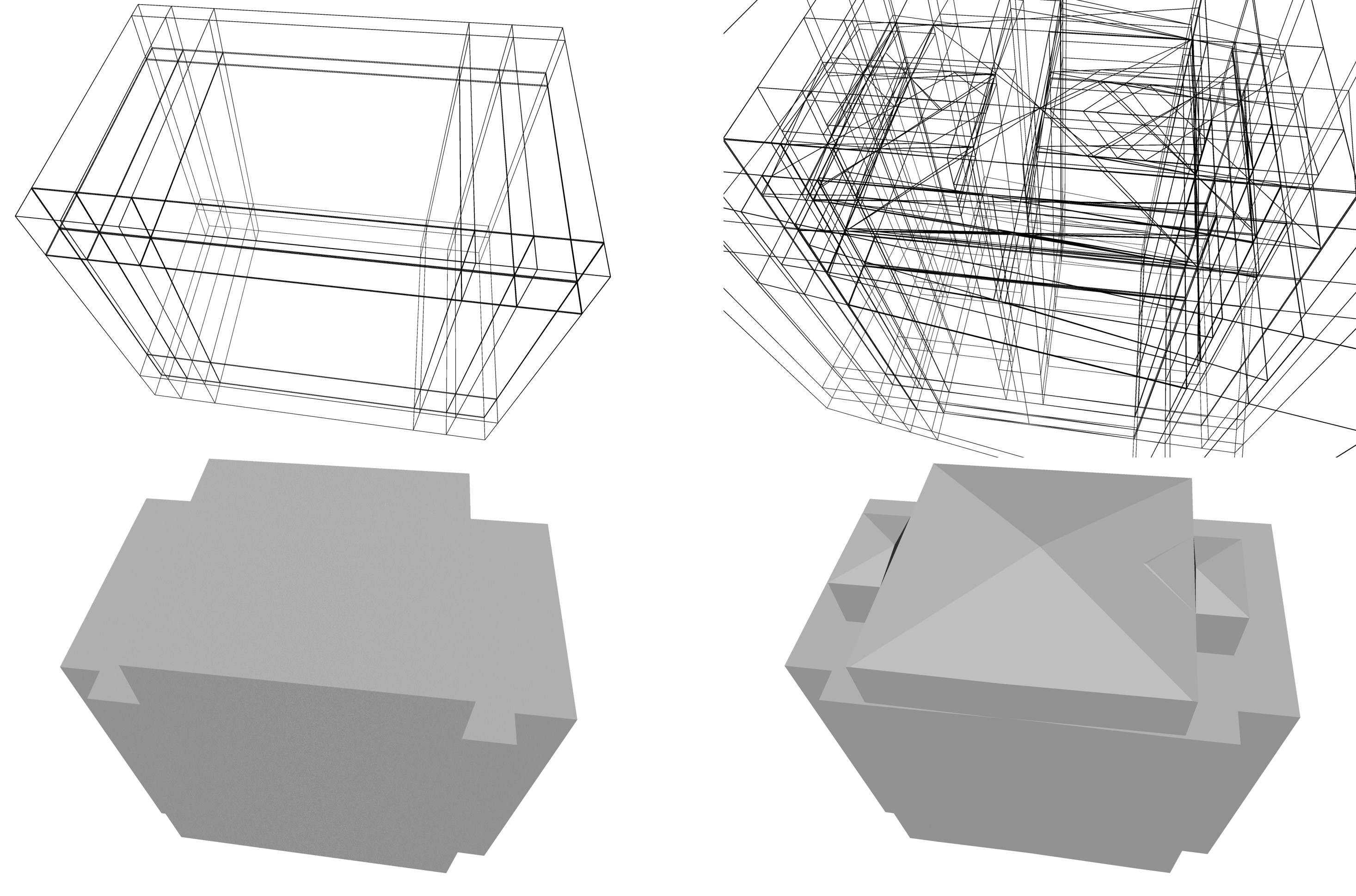,width=0.85\linewidth}}
\caption{Comparison between Manhattan reconstruction~\citep{li2016manhattan} and PolyGNN. Left: Manhattan reconstruction with polycubes as candidates. Right: PolyGNN (ours) with arbitrary polyhedra as candidates.}
\label{fig:comparison_manhattan}
\end{figure}

\begin{table*}[htb]
\begin{tabular}{ccccccc}
\hline
Method                                  & Learning  & $S$ (\%) $\uparrow$ & $F_N$ $\downarrow$ & $H$ (m) $\downarrow$ & $H$ (\%) $\downarrow$  \\ \hline
2.5D DC~\citep{zhou20102}               & \xmark   & \textbf{100.00} &  613.63         & -             & -            \\
City3D~\citep{huang2022city3d}          & \xmark   & 98.86           &  48.87          & 1.10          & 6.0          \\
Geoflow~\citep{peters2022automated}     & \xmark   & 99.97           &  137.66         & 0.42          & 2.8          \\
Points2Poly~\citep{chen2022points2poly} & \cmark   & 99.76           &  \textbf{27.94} & 0.83          & 4.7          \\ \hline
PolyGNN (ours w/ plain)                 & \cmark   & 99.76           &  36.37          & 0.81          & 4.7          \\ 
PolyGNN (ours w/ conv.)                 & \cmark   & \textbf{100.00} &  28.82          & \textbf{0.33} & \textbf{2.2} \\ \hline
\end{tabular}
\caption{Quantitative performance comparison with state-of-the-art methods. ``Learning'' indicates learning-based methods. $S$, $F_N$, and $H$ denote success rate, number of faces, and Hausdorff distance, respectively. Statistics were derived from 10,000 held-out samples, while $F_N$ was derived from successfully reconstructed samples.}
\label{tab:accuracy_comparison}
\end{table*}

\autoref{tab:accuracy_comparison} presents a quantitative comparison between our method and state-of-the-art methods in urban reconstruction, while \autoref{fig:comparison} showcases examples for qualitative comparison as well. The 2.5D DC method~\citep{zhou20102} outputs only facades and roofs, and therefore cannot be fairly compared to other reconstructions quantitatively by Hausdorff distance. Nevertheless, it was unable to represent building models with a concise set of parameters. In contrast, all the other methods exhibit compact reconstructions. Compared to the traditional optimization-based approach City3D~\citep{huang2022city3d}, our method demonstrates the capability to handle more complex buildings commonly found in large-scale urban scenes. City3D adopts exhaustive partitioning, leading to a large solution space for exploration. Thus, it only managed to reconstruct 9,886 buildings within a 5-minute timeout using its Gurobi solver \citep{gurobi}, resulting in inferior reconstruction accuracy as measured by the balanced Hausdorff distance. In contrast, our approach utilizes adaptive space partitioning, resulting in a more compact candidate space that enhances both efficiency and overall accuracy. Compared to Geoflow~\citep{peters2022automated}, which explicitly confines reconstruction within the footprint, our method demonstrates superior performance in terms of accuracy and compactness, even though it is not restricted to the 2.5D disk topology. Furthermore, in comparison to the learning-based method Points2Poly~\citep{chen2022points2poly}, PolyGNN excels in efficiency while achieving comparable or higher geometric accuracy. \autoref{fig:transferability} also demonstrates that PolyGNN transfers better than Points2Poly from synthetic to real data, with lower RMSE.

\begin{figure}[htb]
  \centering
  \centerline{\epsfig{figure=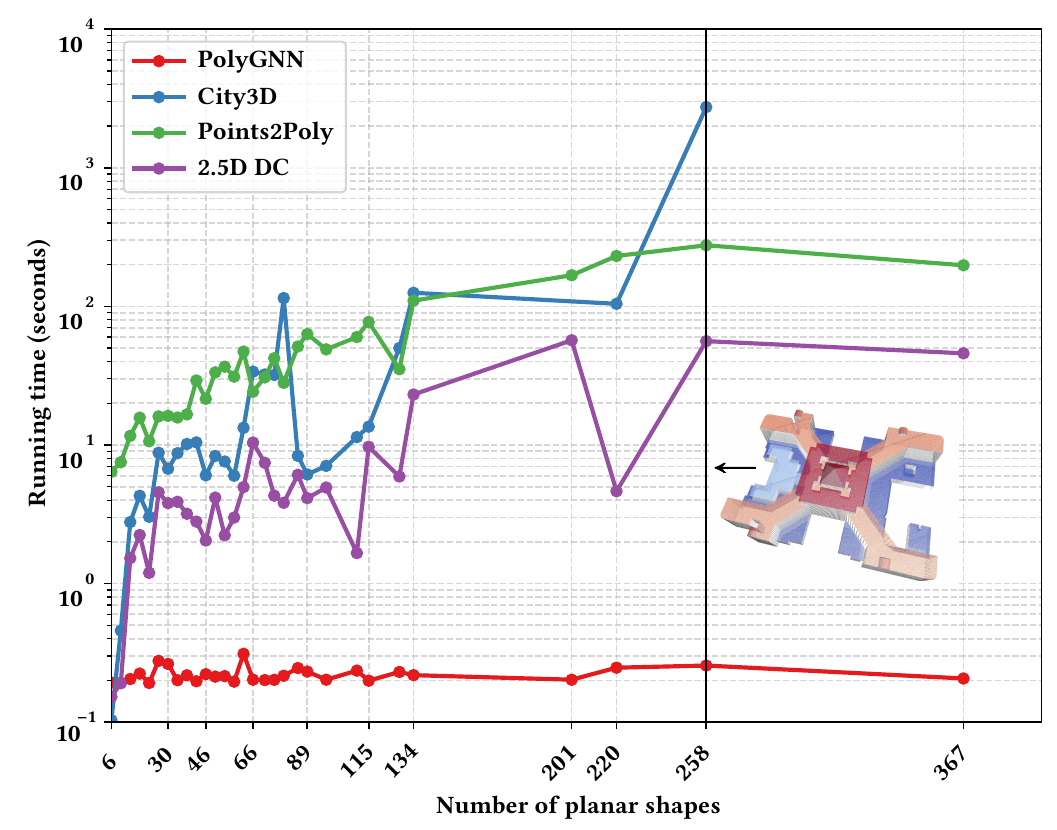,width=0.99\linewidth}}
\caption{Running time comparison. Statistics are derived from 30 buildings of various geometric complexity with the number of planar primitives ranging from 6 to 367.}
\label{fig:runtime_comparison}
\end{figure}

Both \autoref{fig:comparison} and \autoref{fig:comparison_ksr} present comparisons with KSR \citep{bauchet2020kinetic}, another primitive assembly method. When provided with identical input primitives, KSR necessitates additional wall points for achieving non-trivial results, along with pre-computed point normals. In contrast, PolyGNN can directly estimate building occupancy from unorganized and incomplete point clouds. Furthermore, \autoref{fig:comparison_simplification} shows a comparison with LowPolyBuildings \citep{gao2022low}, a simplification approach operating on dense smooth surfaces. Notably, PolyGNN's reconstruction exhibits greater regularity with the same level of complexity. Additionally, \autoref{fig:comparison_manhattan} showcases a comparison with Manhattan reconstruction~\citep{li2016manhattan}, a model-based approach that utilizes polycubes as the model library. Here, PolyGNN's reconstruction demonstrates superior flexibility in describing arbitrarily oriented building geometry.

\begin{table*}[htb]
\begin{tabular}{cccccc}
\hline
Method                                   & Label type     & \#Queries \textit{train}     & \#Queries \textit{test}    & Efficiency \textit{train} & Efficiency \textit{test}  \\ \hline
Points2Poly w/ exh.                      & Class + value  & 2,600,000        &  14,146,600       & 1x          & 1x           \\ 
Points2Poly w/ ada.                      & Class + value  & 2,600,000        &  1,127,100        & 1x          & 13x          \\ \hline
PolyGNN (ours) w/ exh.                   & Class          & 174,112          &  174,112          & 15x         & 81x          \\
PolyGNN (ours) w/ ada.                   & Class          & 13,872           &  13,872           & \textbf{187x}        & \textbf{1020x}        \\ \hline
\end{tabular}
\caption{Efficiency comparison between Points2Poly~\citep{chen2022points2poly} and ours, two learning-based methods. The building in \autoref{fig:architecture} with 60 planar segments is taken for calculating the number of queries. Efficiency is a derived factor
 based on the number of queries; the actual gain may deviate due to parallelization.}
\label{tab:comparison_points2poly}
\end{table*}

\autoref{fig:runtime_comparison} presents the running time comparison among different methods, highlighting the superior efficiency of our approach. City3D, which relies on an integer programming solver, encounters computational bottlenecks as the number of planar primitives increases. Consequently, for certain complex buildings, the reconstruction cannot even be solved within a feasible time frame of 24 hours. Points2Poly requires approximately 4 days for one epoch of training, whereas PolyGNN only takes 24 minutes (240$\times$ faster). The longer inference time of Points2Poly, on the other hand, comes mostly from two factors. Firstly, more efforts are required for its occupancy estimation. \autoref{tab:comparison_points2poly} presents the comparison with the learning-based method Points2Poly, for reconstructing the building in \autoref{fig:architecture} with 60 planar segments. Points2Poly enumerates queries with signed distance values to learn a smooth boundary, while ours only requires discrete binary-class queries directly describing the piecewise planar surface. Meanwhile, the adaptive strategy significantly reduces the number of queries for both training and testing of our method as fewer polyhedra need to be considered as candidates. Secondly, the interface computation, which is necessary for assigning graph edge weights, contributes to the longer running time of Points2Poly. In contrast, PolyGNN can reconstruct a building directly by inferring the polyhedral occupancy, leveraging GPU parallelization for improved efficiency.

\subsection{Robustness analysis}

\begin{figure}[htb]
  \centering
  \centerline{\epsfig{figure=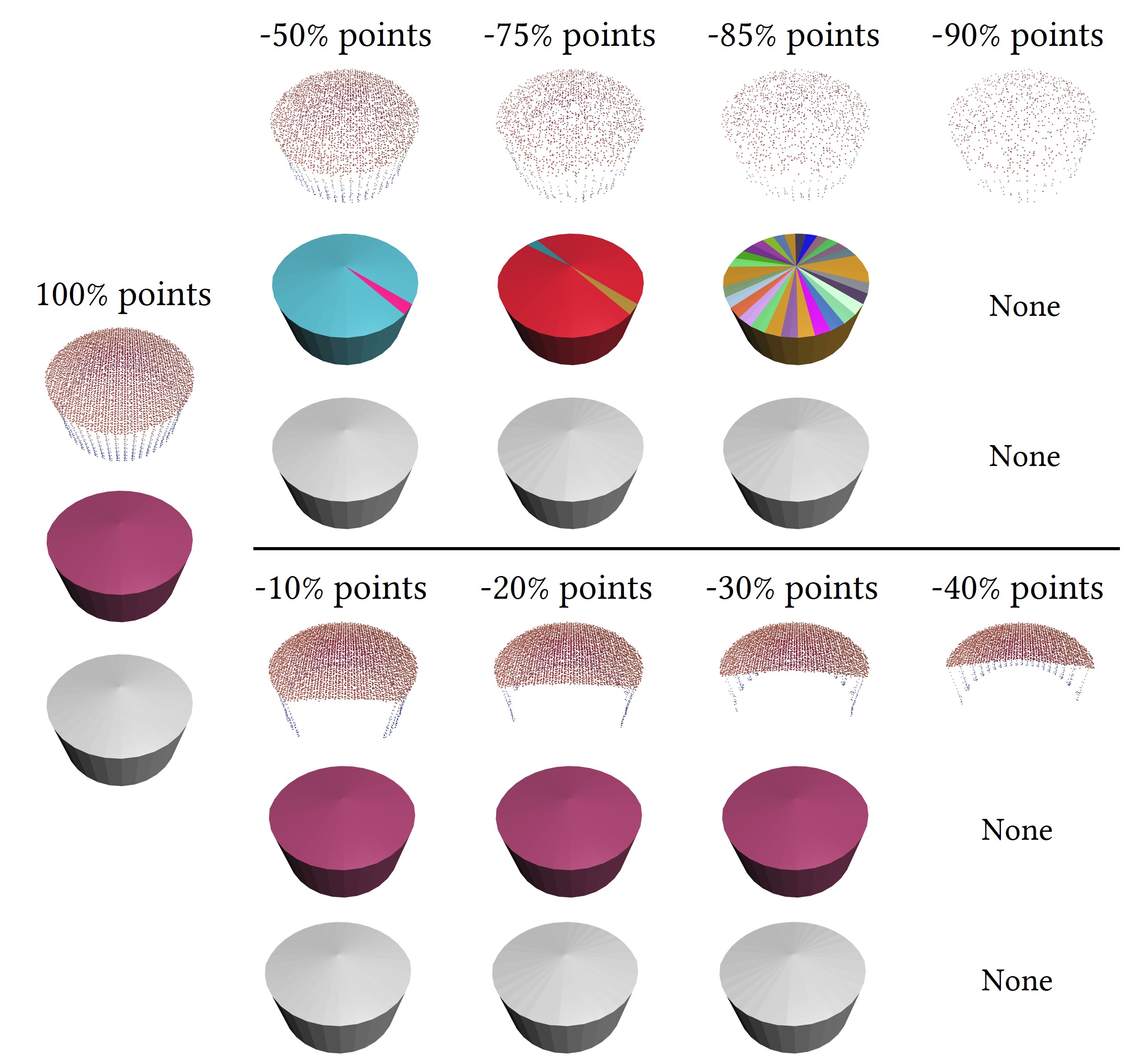,width=0.99\linewidth}}
\caption{Robustness to point cloud distribution with various point densities (top) and levels of missing points (bottom). Each triplet from top to bottom: input point cloud, polyhedra classified as building components, and reconstructed model. Polyhedra are randomly color-coded.}
\label{fig:robustness_distribution}
\end{figure}

To analyze the robustness of PolyGNN against variations in point cloud density, we randomly drop input points. As shown in \autoref{fig:robustness_distribution}, even though PolyGNN was not explicitly trained against various point densities, it manages to achieve reasonable reconstruction with sparsely subsampled point clouds. When uniformly dropping 85\% of the points, the overall shape can still be maintained as long as planar primitives remain accurate. Additionally, while preserving the decomposition, we truncated different portions of input points to analyze the robustness of PolyGNN against heterogeneously missing points, a scenario for which it was also not explicitly trained. The reconstruction remains feasible even when up to 30\% of the points are truncated, as shown in \autoref{fig:robustness_distribution}.

\begin{figure}[htb]
  \centering
  \centerline{\epsfig{figure=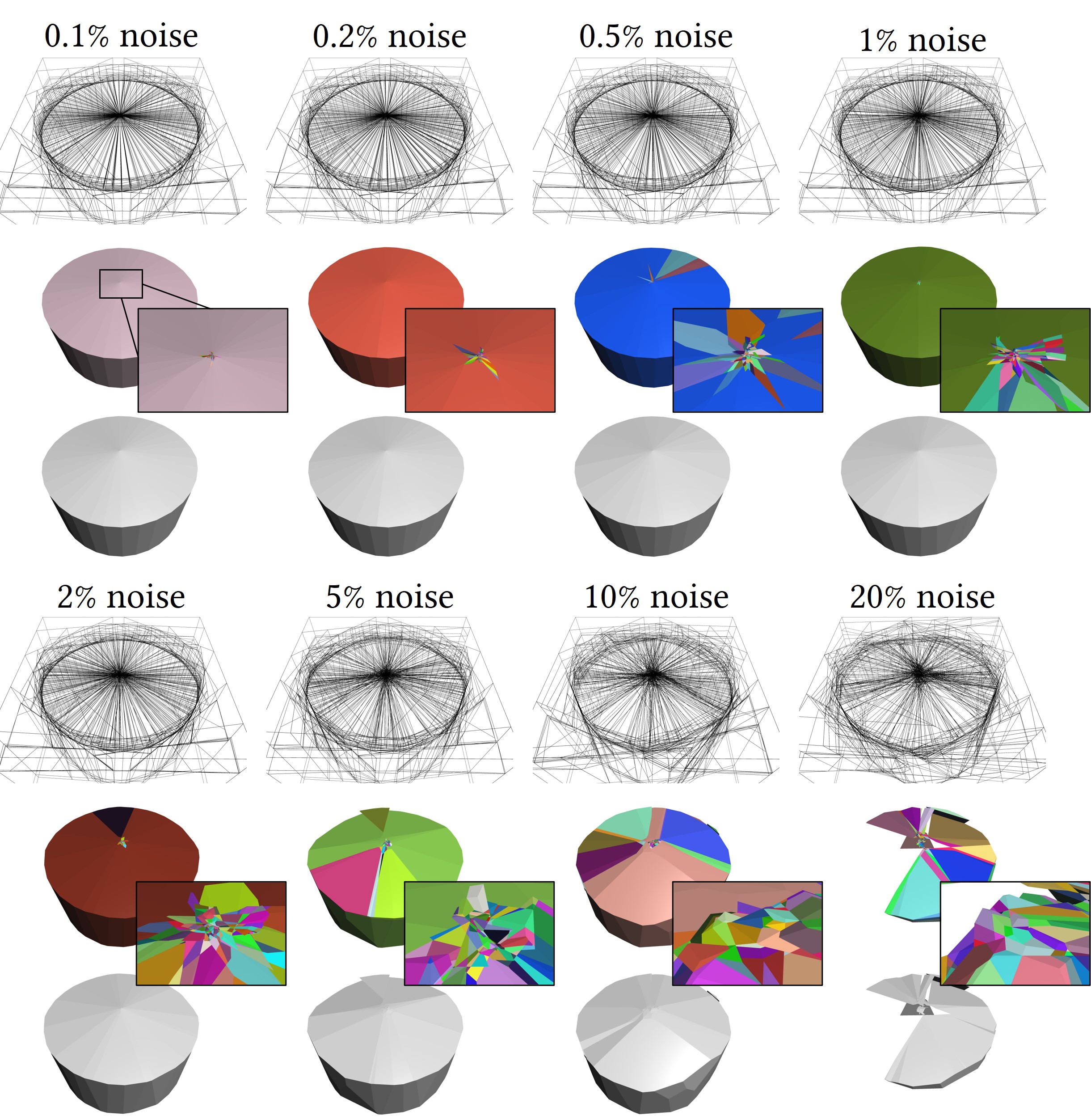,width=0.99\linewidth}}
\caption{Robustness to quality of planar primitives. Increasing levels of Gaussian noise are added to perturb the initial planar primitives. From top to bottom: cell complex, polyhedra classified as building components, and reconstructed model. Polyhedra are randomly color-coded.}
\label{fig:robustness_normal}
\end{figure}

We also apply different levels of noise to perturb the initial planes. As shown in \autoref{fig:robustness_normal}, the reconstruction starts to deviate from the ground truth when the noise level reaches 5\%. Increasing the noise level results in irregular cell complexes and thus poses challenges for primitive assembly. With 20\% noise, the method fails to reconstruct the main body. Since PolyGNN learns a discrete decision boundary, it is inherently sensitive to the quality of planar primitives. We believe that introducing different point distributions and noises into the training data would further enhance the model's robustness.

\subsection{Limitations}

We assume the availability of high-quality planar primitives extracted from point clouds. This assumption may not always be fulfilled with real-world data that contains significant noise and occlusions, and therefore is considered a limitation of the proposed method and similar approaches that rely on primitive assembly. Additionally, since PolyGNN operates on individual buildings, a preliminary step of building instance segmentation is required prior to reconstruction, especially when dealing with point clouds of entire scenes.

\begin{figure}[t]
  \centering
  \centerline{\epsfig{figure=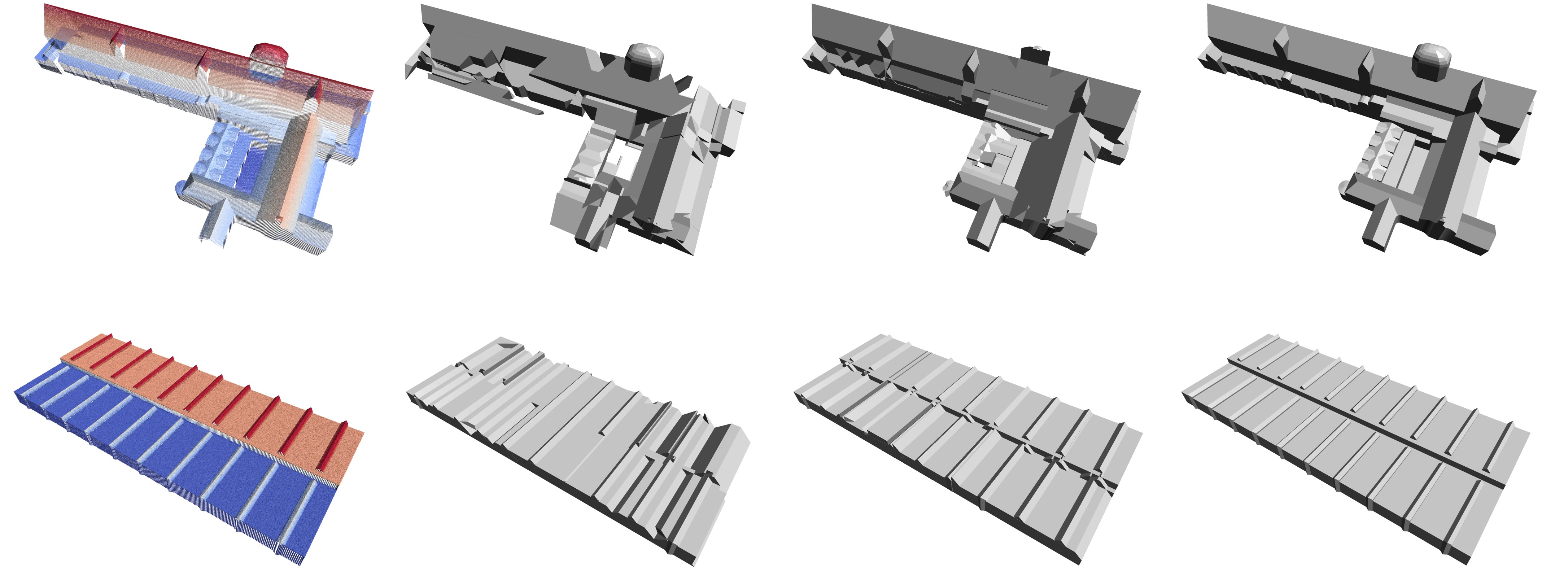,width=0.95\linewidth}}
\caption{Suboptimal reconstruction of buildings with complex structures. From left to right: input point cloud color-coded by height field, reconstructed model with the plain encoder, reconstructed model with the convolutional encoder, and ground truth.}
\label{fig:failure}
\end{figure}

\autoref{fig:failure} shows some failure cases. When reconstructing buildings with complex structures, PolyGNN may encounter challenges in capturing fine details, such as intricate rooftop superstructures. These failures can be attributed to two factors. Firstly, the complexity of a building implies a larger and more intricate polyhedral embedding, which poses challenges to the network's prediction. Secondly, the training dataset predominantly consists of buildings with simple shapes, leading to an under-representation of complex structures. As a result, the network may have limited exposure to and understanding of complex architectural elements with fine details.

\section{Conclusion}

We introduced PolyGNN, a novel framework for urban building reconstruction with a polyhedron-based graph neural network. Unlike traditional deep implicit fields that learn a continuous function, our approach learns a piecewise planar occupancy function derived from polyhedral decomposition. We proposed a skeleton-based sampling strategy for representing an arbitrary-shaped polyhedron within the neural network, and demonstrated its superior performance compared to other variants. Furthermore, PolyGNN is end-to-end optimizable and is designed to accommodate variable-size input points, polyhedra, and queries with an index-driven batching technique. 

We developed PolyGNN on a large-scale synthetic building dataset furnished with polyhedral labels and analyzed its transferability on cross-city synthetic data and real-world data. Both qualitative and quantitative results demonstrate the effectiveness of PolyGNN, particularly in terms of efficiency. Moreover, our framework is designed to be generic. It can potentially be extended to handle other types of point clouds, such as photogrammetric ones, and can be utilized for the reconstruction of generic piecewise planar 3D objects beyond buildings.

Finally, we remark on the gap between synthetic and real-world point clouds. In future work, we aim to bridge this gap further, enabling learning-based reconstruction methods to better abstract and leverage a vast volume of training data. Additionally, we intend to explore techniques for integrating semantic attributes to enrich the polyhedral graph and integrate plane extraction into the neural architecture.

\section*{Acknowledgement}
This work is jointly funded by the TUM Georg Nemetschek Institute for Artificial Intelligence for the Built World as part of the AI4TWINNING project, by the German Federal Ministry of Education and Research (BMBF) in the framework of the international future AI lab ``AI4EO -- Artificial Intelligence for Earth Observation: Reasoning, Uncertainties, Ethics and Beyond'' (grant number: 01DD20001), by the German Federal Ministry for the Environment, Nature Conservation, Nuclear Safety and Consumer Protection (BMUV) based on a resolution of the German Bundestag (grant number: 67KI32002B; Acronym: \textit{EKAPEx}) and by Munich Center for Machine Learning.

\bibliographystyle{cas-refs}

\bibliography{main}

\end{document}